\documentclass[10pt,journal]{IEEEtran}
\usepackage{amsmath,amsfonts}
\usepackage{algorithmic}
\usepackage{algorithm}
\usepackage{array}
\usepackage[caption=false,font=normalsize,labelfont=sf,textfont=sf]{subfig}
\usepackage{textcomp}
\usepackage{stfloats}
\usepackage{cite}
\usepackage{url}
\usepackage{booktabs}
\usepackage{verbatim}
\usepackage{graphicx}
\usepackage{cite}
\usepackage{xcolor}
\usepackage{multirow}

\begin{document}

\title{Interpretable Hypergraph Learning via \\ Neural Additive Models}

\author{Shihan Feng, Xin Zheng, Shiyi Yang, Ren Wang, Chudi Zhong, and Can Chen
\IEEEcompsocitemizethanks{\IEEEcompsocthanksitem Shihan Feng and Xin Zheng contributed equally to this work. Corresponding authors: Ren Wang, Chudi Zhong, Can Chen.}
\IEEEcompsocitemizethanks{\IEEEcompsocthanksitem Shihan Feng and Xin Zheng are with the School of Data and Information Sciences, University of North Carolina at Chapel Hill, Chapel Hill, NC 27599, USA (e-mail: shihanf@unc.edu; xinzheng@unc.edu).}
\IEEEcompsocitemizethanks{\IEEEcompsocthanksitem Shiyi Yang is with the Department of Biostatistics, University of North Carolina at Chapel Hill, Chapel Hill, NC 27599, USA (e-mail: shiyi@email.unc.edu).}
\IEEEcompsocitemizethanks{\IEEEcompsocthanksitem Ren Wang is with the Department of Electrical and Computer Engineering, Illinois Institute of Technology, Chicago, IL 60616, USA (e-mail: rwang74@illinoistech.edu).}
\IEEEcompsocitemizethanks{\IEEEcompsocthanksitem Chudi Zhong is with the School of Data and Information Sciences and the Department of Statistics and Operations Research, University of North Carolina at Chapel Hill, Chapel Hill, NC 27599, USA (e-mail: chudi@unc.edu).}
\IEEEcompsocitemizethanks{\IEEEcompsocthanksitem Can Chen is with the School of Data and Information Sciences, the Department of Mathematics, and the Department of Biostatistics, University of North Carolina at Chapel Hill, Chapel Hill, NC 27599, USA (e-mail:canc@unc.edu).}
}

\maketitle

\begin{abstract}
Hypergraphs offer a natural framework for modeling  networked data, where dependencies among entities are governed by higher-order interactions. While hypergraph learning methods such as hypergraph neural networks have demonstrated remarkable predictive performance, most existing approaches rely on black-box message-passing architectures, making it difficult to disentangle the contributions of node attributes and higher-order structural information. To address this challenge, we introduce the hypergraph neural additive network (HGNAN), an inherently interpretable framework for learning on hypergraph-structured data. HGNAN extends classical neural additive models to higher-order relational data by integrating feature-wise nonlinear decomposition with hypergraph-aware structural aggregation, enabling transparent prediction for both node- and hyperedge-level tasks. Extensive experiments on  benchmark  datasets demonstrate that HGNAN achieves performance comparable with state-of-the-art hypergraph learning methods while providing intrinsic and meaningful interpretability.
\end{abstract}

\begin{IEEEkeywords}
Hypergraphs, hypergraph learning, network representation learning, interpretable machine learning.
\end{IEEEkeywords}

\section{Introduction}\label{introduction}
\IEEEPARstart{I}{nteractions} in many empirical systems extend beyond pairs of entities and instead occur among groups~\cite{BATTISTON20201}. Such higher-order interactions arise naturally in social networks~\cite{benson2018simplicial, cencetti2021temporal, zhu2018social}, cellular networks~\cite{klamt2009hypergraphs}, and cyber-physical networks~\cite{zhang2019introducing}. Hypergraphs provide a direct representation of these systems by allowing each hyperedge to connect an arbitrary number of nodes \cite{chen2020tensor,chen2021controllability}.
Through message passing over this higher-order structure, hypergraph neural networks (HGNNs) learn predictive node embeddings by integrating node and hyperedge attributes~\cite{gao2020hypergraph}. HGNNs support both node-level prediction, such as cell-type identification~\cite{li2023scmhnn}, gene disease-risk prediction~\cite{ma2025hypergraph}, and cancer-subtype classification~\cite{wang2024hypertmo}, and hyperedge-level prediction, such as  metabolic reaction recovery \cite{NHP,chen2023teasing}, inference of genomic relationships~\cite{dotson2022deciphering, su2025inferring}, and drug interaction prediction~\cite{nguyen2022sparse,wu2025heterogeneous}. In scientific applications, however, accurate prediction is often only the first step. A predicted cell type, disease-associated gene, or regulatory relationship may constitute a scientific hypothesis whose validity and underlying mechanism require further investigation. Consequently, it is important to understand what drives an HGNN prediction: the attributes of the entities involved, their organization within the higher-order network structure, or an interplay between the two.

Existing HGNNs are primarily designed for predictive performance rather than separating these sources of information. Representative architectures include spectral or Laplacian-based methods~\cite{ijcai2021p353,Fiorini2024LetTB}, graph-reduction or expansion-based methods~\cite{Chen2020,NEURIPS2019_1efa39bc}, set-function-based methods~\cite{Chien2021YouAA}, and more recent expressive or efficient variants~\cite{wang2022equivariant,wang2023hypergraph,10.1609/aaai.v39i20.35472,bae2026parameter}. Despite these architectural differences, they all apply feature transformations and structural aggregation jointly inside each layer. As a result, the two contributions are entangled and can not be explictly identifiable in the final prediction. To address this lack of transparency, post-hoc explainers attempt to separate contributions after training. GNNExplainer~\cite{NEURIPS2019_d80b7040}, PGExplainer~\cite{PGExplainer}, and XGNN~\cite{XGNN} identify subgraphs or features that most influence a trained graph neural network (GNN), and HyperEX~\cite{maleki2023learning} and SHypX~\cite{Su2024ExplainingHN} extend this paradigm to HGNNs by extracting explanatory sub-hypergraphs. While effective in many settings, such post-hoc explanations do not constrain the underlying model itself and their explanations are not guaranteed to faithfully reflect its true reasoning process, which can undermine trust in high-stakes applications \cite{RudinDontExplain}.

An alternative direction is to design models that are intrinsically interpretable. Additive models provide a natural realization of this principle. In a generalized additive model (GAM)~\cite{hastie1986generalized}, the prediction is expressed as a sum of feature-specific univariate shape functions, so each fitted function directly represents that feature's contribution. Neural additive models (NAMs) \cite{NEURIPS2021_251bd044} extend this idea by learning a small neural network to model each nonlinear feature effect. For relational data, however, interpretation must additionally account for structure. Graph neural additive networks (GNANs)~\cite{bechlerspeicher2024} incorporate pairwise structure by weighting neighboring feature responses according to graph distance. However,  this construction does not directly extend to hypergraphs.  Although a hypergraph can be transformed into a graph, different reductions, such as clique and intersection expansions, induce different neighborhoods and distance relationships. Consequently, an additive model applied to such a reduction therefore attributes effects to the chosen graph representation rather than to the original higher-order structure. Existing interpretable-by-design hypergraph models do not resolve this issue: SHy~\cite{yu2025self}, for example, explains diagnostic predictions through extracted phenotypes but does not separate attribute effects from higher-order structural effects. These limitations motivate two fundamental questions: \textit{Can an additive predictor separate feature and structural effects directly on a hypergraph, and can it do so without sacrificing predictive accuracy or attribution faithfulness?}

In this paper, we introduce the hypergraph neural additive network (HGNAN), the first neural additive architecture specifically designed for hypergraph learning. HGNAN extends NAMs from independent tabular inputs to hypergraph-structured data. Rather than relying on post-hoc explanations, HGNAN makes the learning algorithm itself inherently decomposable by expressing predictions as the sum of feature-wise nonlinear responses and explicit hypergraph-induced structural influences. Its two variants, HGNAN-node and HGNAN-edge, provide decompositions at the levels of attributes, neighboring entities, and structural distance for node- and hyperedge-level prediction, respectively. Our empirical results show that this inherently decomposable formulation achieves predictive performance competitive with both classical and state-of-the-art HGNN architectures while providing faithful attributions across diverse hypergraph datasets.

The remainder of this paper is organized as follows. Section~\ref{preliminaries} introduces hypergraph notations and hypergraph distances. Section~\ref{methods} presents the formulations of HGNAN-node and HGNAN-edge, together with their training procedures.  Section~\ref{experiments} evaluates HGNAN-node against nine HGNN baselines on node classification and HGNAN-edge against three baselines on hyperedge prediction, and further presents ablation and faithfulness analyses as well as two case studies comparing the learned decompositions with domain knowledge. Section~\ref{conclusion} concludes the paper with future directions.

\section{Preliminaries}\label{preliminaries}
\subsection{Hypergraphs}
We consider an undirected hypergraph \( \mathcal{H} = (\mathcal{V}, \mathcal{E}) \), where \( \mathcal{V}=\{v_1,v_2,\dots,v_n\} \) is a set of \(n\) nodes and \( \mathcal{E}=\{e_1,e_2,\dots,e_m\} \) is a set of \(m\) hyperedges. Each hyperedge \(e_j\in\mathcal{E}\) is a non-empty subset of nodes, i.e., \(e_j\subseteq\mathcal{V}\).
The hypergraph structure can be represented by an incidence matrix \( \mathbf{H}\in\{0,1\}^{n\times m} \), where \( \mathbf{H}_{ij}=1 \) if node \(v_i\) is contained in hyperedge \(e_j\), and \( \mathbf{H}_{ij}=0 \) otherwise.
Each node \(v_i\) is associated with a feature vector \( \mathbf{x}_i\in\mathbb{R}^p \), and the corresponding node feature matrix is \( \mathbf{X}\in\mathbb{R}^{n\times p} \). Similary, each hyperedge $e_i$ can be optionally associated with a feature vector $\mathbf{y}_i \in \mathbb{R}^{p_e}$, and the corresponding hyperedge feature matrix is  \( \mathbf{Y}\in\mathbb{R}^{m\times p_e} \)

\subsection{Distance on Hypergraphs}
\label{sec:distance_on_hypergraphs}
To characterize structural connectivity at different scales, we define distances through the overlap between hyperedges. For an overlap threshold $s\in\mathbb{Z}^+$, the $s$-intersection graph $G^{(s)}$ has one vertex for each hyperedge in $\mathcal{E}$. Two distinct hyperedges are adjacent if they share at least $s$ nodes. Accordingly, its  adjacency matrix $\mathbf{A}^{(s)}\in\{0,1\}^{m\times m}$ is  defined as
\begin{equation}
    \mathbf{A}^{(s)}_{ij}
    =
    \mathbf{1}\!\left[i\neq j\ \text{and} \ |e_i\cap e_j|\ge s\right].
\end{equation}
Varying $s$ captures different scales of higher-order connectivity: smaller values induce broader connectivity among hyperedges, whereas larger values identify stronger structural relations through more stringent overlap requirements.

For any pair of hyperedges \(e_i,e_j\in\mathcal{E}\), we define \(d^{(s)}(e_i,e_j)\) as the shortest-path distance between \(e_i\) and \(e_j\) in the \(s\)-intersection graph \(G^{(s)}\), with
\(d^{(s)}(e_i,e_i)=0.\) If \(e_i\) and \(e_j\) are disconnected in \(G^{(s)}\), we set \(d^{(s)}(e_i,e_j)=\infty\).
Thus, \(d^{(s)}(e_i,e_j)\) measures the minimum number of \(s\)-adjacent hyperedge hops needed to connect \(e_i\) and \(e_j\).
We also define an induced node distance from the hyperedge distance.
For two distinct nodes \(v_i,v_j\in\mathcal{V}\), define
\begin{equation}
    d^{(s)}(v_i,v_j)
    =
    1+
    \min_{\substack{e_p,e_q\in\mathcal{E}:\\ v_i\in e_p,\ v_j\in e_q}}
    d^{(s)}(e_p,e_q),
    \label{eq:node_distance}
\end{equation}
and we set the self-distance to \(d^{(s)}(v_i,v_i)=0\). If no connected pair of hyperedges containing \(v_i\) and \(v_j\) exists, we set \(d^{(s)}(v_i,v_j)=\infty\). Under this definition, a node is at distance \(0\) from itself, two distinct nodes that appear in the same hyperedge are at distance \(1\), and nodes whose containing hyperedges are adjacent in \(G^{(s)}\) are at distance \(2\). Therefore, the \(s\)-neighborhood of node \(v_i\) can be defined as
\begin{equation}
    \mathcal{N}_i^{(s)}
    =
    \{v_j\in\mathcal{V}: d^{(s)}(v_i,v_j)\le 2\}.
\end{equation}
This neighborhood contains the target node \(v_i\) itself, the nodes on the same hyperedge as \(v_i\), and the nodes connected through one \(s\)-adjacent hyperedge hop. 

To normalize contributions from hyperedges at the same distance, we use distance-shell normalization. For a target hyperedge \(e_i\), define the hyperedge distance shell and shell cardinality as
\begin{equation}
    \mathcal{S}^{(s)}_{i}(r)
    =
    \{e_j\in\mathcal{E}: d^{(s)}(e_i,e_j)=r\},
\end{equation}
\begin{equation}
    \# d^{(s)}(e_i,e_j)
    =
    \left|\mathcal{S}^{(s)}_{i}(d^{(s)}(e_i,e_j))\right|.
\end{equation}
The shell cardinality prevents distances with many hyperedges from dominating the aggregation solely due to their shell size.
Finally, we use  \(s_{\max}\) to denote the largest overlap threshold  and consider  \(s\in\{1,2,\dots,s_{\max}\}\).

\section{Methods} \label{methods}
\begin{figure*}[ht]
    \centering
    \includegraphics[width=0.95\linewidth]{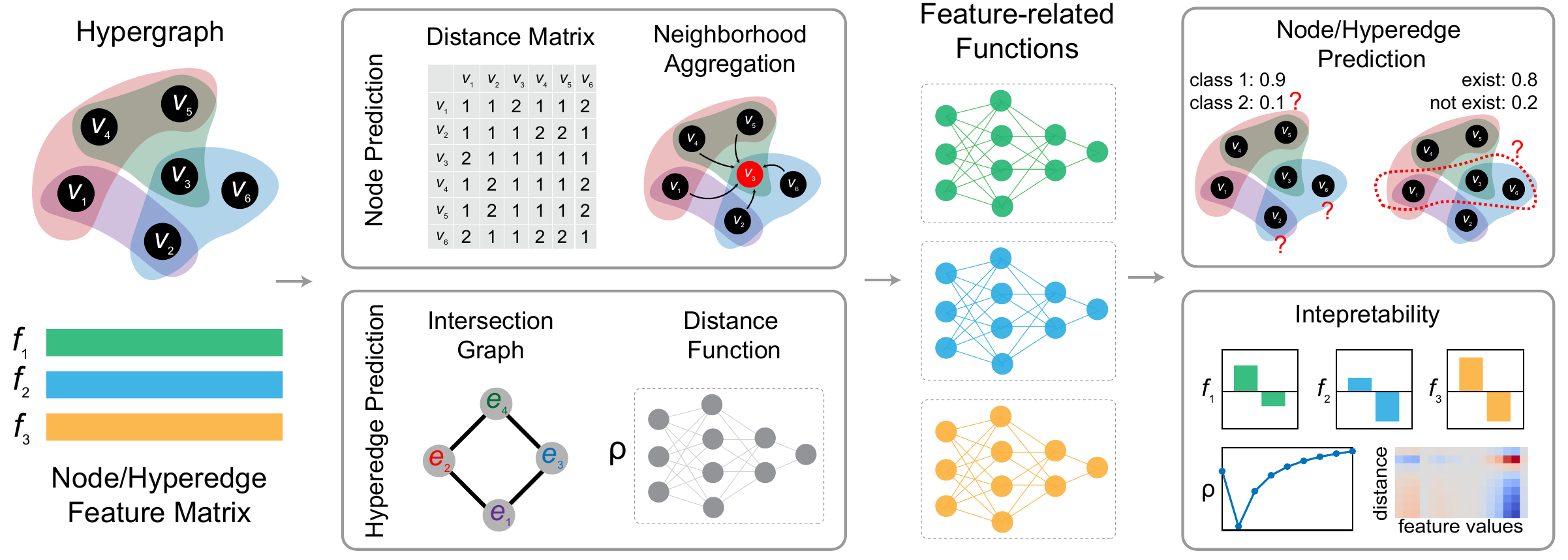}
    \caption{
    Overview of HGNAN.
    HGNAN takes a hypergraph and a feature matrix as input and processes them through two interpretable components: feature-wise nonlinear transformations and structure-aware influence functions.
    For node prediction, HGNAN performs neighborhood aggregation over hypergraph-structured information using $s$-adjacency.
    For hyperedge prediction, it builds $s$-intersection graph and applies a distance-aware function to capture structural influence between hyperedges.
    As in NAMs, each feature-related function operates only on a single feature, which preserves feature-level decomposability.
    }
    \label{fig:overview}
\end{figure*}

In this section, we present hypergraph neural additive network (HGNAN), an inherently interpretable HGNN architecture for learning on hypergraph-structured data (see Fig.~\ref{fig:overview}).
HGNAN learns feature-wise effects through a NAM-style additive structure, while incorporating hypergraph structural information via either local neighborhood aggregation or global distance-based modeling. Accordingly, HGNAN consists of two interpretable components: feature shape functions, which model how individual input features contribute to predictions, and structural functions, which model how higher-order relations in  hypergraph modulate these contributions. Notably, HGNAN supports both node-level and hyperedge-level prediction through two instantiations, termed HGNAN-node and HGNAN-edge, respectively. The key distinction between these two instantiations is how they incorporate hypergraph structure information. HGNAN-node performs node classification by aggregating information over local neighborhoods defined by the hypergraph's higher-order adjacency structure, whereas HGNAN-edge learns a distance-based function over a set of higher-order intersection graphs to model structural influence among hyperedges. Both variants preserve the NAM-style feature-wise decomposition while extending it to relational structures that arise naturally in hypergraphs.

\subsection{Node-Level Prediction}
\label{sec:node_pred}

HGNAN-node generalizes NAM-style additive modeling to node-level prediction on hypergraphs. Given node features \(\mathbf{x}_i\in\mathbb{R}^p\), HGNAN-node learns a set of feature-wise shape functions \(\{f_k\}_{k=1}^p\), where each \(f_k:\mathbb{R}\to\mathbb{R}\) transforms the \(k^{\mathrm{th}}\) feature independently. This preserves the additive structure of NAMs while allowing each feature contribution to be aggregated over a hypergraph-induced neighborhood. We first consider binary classification, for which each feature shape function produces a scalar output.  For a node \(v_i\) and adjacency order \(s\), HGNAN-node computes the \(k^{\mathrm{th}}\) feature-channel contribution as
\begin{equation}
    [\mathbf{h}_i^{(s)}]_k=\sum_{v_j\in\mathcal{N}_i^{(s)}}a_{ij}^{(s)}f_k([\mathbf{x}_j]_k),
    \label{eq:hgnan_node_feature_channel}
\end{equation}
where \(\mathcal{N}_i^{(s)}\) is the \(s\)-neighborhood of \(v_i\) and \(a_{ij}^{(s)}\) is a learned neighbor weight on $G^{(s)}$. The same structural weight \(a_{ij}^{(s)}\) is shared across feature channels, so it represents the structural contribution of neighbor \(v_j\) to the prediction of \(v_i\), independent of any particular feature.
The neighbor weight \(a_{ij}^{(s)}\) is computed by a GAT-style small neural network from node representations~\cite{VelickovicCCRLB18}.
In practice, we impose a sparse constraint to \(a_{ij}^{(s)}\) so that only a small subset of neighbors receives non-negligible weights. After training, we compute and cache the attention weights and reuse them without recomputation during inference. Therefore, the strict additivity preserves and it can provide neighbor-level explanation for node prediction, where larger \(a_{ij}^{(s)}\) indicates that \(v_j\) contributes more strongly to the prediction of \(v_i\) under \(s\)-adjacency.

After obtaining feature-channel contributions, HGNAN-node combines them using a learnable feature weight \(w_k\in\mathbf{w}\),
\begin{equation}
    z_i^{(s)}
    =
    \sum_{k=1}^{p}
    w_k [\mathbf{h}_i^{(s)}]_k,
    \label{eq:hgnan_node_feature_gate}
\end{equation}
The weight vector \(\mathbf{w}\in\mathbb{R}^p\) is only dependent on feature $k$ and therefore shared across all classes. We normalize \(\mathbf{w}\) by a softmax so that \(w_k\ge 0\) and \(\sum_{k=1}^p w_k=1\), allowing \(w_k f_k(\cdot)\) to be interpreted as a feature-level contribution. The scalar \(z_i^{(s)}\) represents the additive score for node \(v_i\) under adjacency order \(s\).
Finally, HGNAN-node aggregates information across different adjacency orders, i.e., 
\begin{equation}
    z_i=\sum_{s=1}^{s_{\max}}
    \beta_s z_i^{(s)},
    \label{eq:hgnan_node_scale_gate}
\end{equation}
where \(\boldsymbol{\beta}\in\mathbb{R}^{s_{\max}}\) is a learnable scale-weight vector.
We normalize \(\boldsymbol{\beta}\) with a softmax over \(s\), so that \(\beta_s\) can be interpreted as the relative contribution of the \(s\)-adjacency scale. Eqs.~\eqref{eq:hgnan_node_feature_channel} and \eqref{eq:hgnan_node_feature_gate} allow HGNAN-node to flexibly learn each neighbor’s contribution to node $i$ by considering both high-order adjacency and feature-level importance. The resulting score \(z_i\) is passed to a linear prediction head to reserve the linear additivity of our model.

In a more generalized \(C\)-class ($C>2$) classification setting, HGNAN-node uses vector-valued shape functions \(f_k:\mathbb{R}\to\mathbb{R}^C\). The same decomposition is applied component-wise, producing class-specific scores \(\mathbf{z}_i^{(s)}, \ \mathbf{z}_i\in\mathbb{R}^C\), where each component corresponds to one class. Overall, HGNAN-node preserves the additive structure of NAMs while extending it to higher-order relational information through explicit feature-, neighbor-, and scale-level contributions. This decomposition makes each node prediction directly attributable to the features and hypergraph neighborhoods that drive it.

\subsection{Hyperedge-Level Prediction}
\label{sec:hyperedge_prediction}
Most existing hyperedge prediction methods first compute node embeddings and then aggregate them through a pooling operation to obtain hyperedge representations. While effective for prediction, this two-stage design can obscure the mechanism behind a hyperedge-level decision as the final prediction is pooled node-level embeddings rather than explicit hyperedge-level features and structural relations among hyperedges. This limitation is particularly important in scientific applications, where one often needs to understand which properties of a candidate hyperedge and which higher-order structural relations support a predicted interaction. Moreover, while local neighborhood aggregation is most effective when label-relevant information is concentrated among nearby neighbors, hyperedge labels can exhibit nonlocal distance-dependent structure, with same-label ratios increasing beyond immediate neighborhoods (see Fig.~\ref{fig:homophily} in Appendix~\ref{dataset_info} for an example). This suggests that informative structural signals may extend beyond local hyperedge neighborhoods. To capture such nonlocal dependencies, we propose HGNAN-edge, which leverages \(s\)-intersection graphs and performs global, distance-aware aggregation for hyperedge prediction.

In HGNAN-edge, we construct initial hyperedge features from the raw node features as $\mathbf{y}_j=\Gamma\big(\{\mathbf{x}_l: v_l\in e_j\}\big)\in \mathbb{R}^{p_e},$ where \(\Gamma(\cdot)\) is a domain-specific feature construction function, and $p_e$ need not equal the node feature dimenion $p$. The choice of $\Gamma$ determines the meaning of each hyperedge feature. For example, in metabolic reaction prediction, hyperedge features can encode changes in functional-group counts before and after a reaction. Learning directly from such explicit hyperedge features therefore enables HGNAN-edge to retain glass-box interpretability at the hyperedge level.

With each hyperedge represented by $\mathbf{y}_j$, hyperedge prediction can be formulated as a learning problem on the \(s\)-intersection graph, whose vertices correspond to hyperedges in the original hypergraph. HGNAN-edge applies the additive modeling principle in this hyperedge domain while preserving the hypergraph-native meaning of distance and structural influence. For a hyperedge \(e_i\) and adjacency order \(s\), define
\begin{equation}
    r_{ij}^{(s)} = \frac{1}{1+d^{(s)}(e_i,e_j)}
\end{equation}
for every hyperedge \(e_j\) with \(d^{(s)}(e_i,e_j)<\infty\). The quantity \(r_{ij}^{(s)}\) is larger for closer hyperedges and is \(1\) when \(e_i=e_j\).

We first present the binary classification case. For a hyperedge $e_j$ and overlap threshold $s$, HGNAN-edge computes the \(k^{\mathrm{th}}\) feature-channel contribution as
\begin{equation}
    [\mathbf{g}_i^{(s)}]_k
    =
    \sum_{j:\, d^{(s)}(e_i,e_j)<\infty}
    \frac{1}{\# d^{(s)}(e_i,e_j)}
    \rho_s\!\left(r_{ij}^{(s)}\right)
    f_k([\mathbf{y}_j]_k),
    \label{eq:hgnan_edge_feature_channel}
\end{equation}
where \(f_k:\mathbb{R}\to\mathbb{R}\) is a learned feature-wise shape function associated with feature \(k\), and \(\rho_s:\mathbb{R}\to\mathbb{R}\) is a learned distance-based modulation function on the \(s\)-intersection graph. The shell normalization term \(\# d^{(s)}(e_i,e_j)\) prevents distance shells containing many hyperedges from dominating the aggregation solely due to their size. Each \(f_k\) and \(\rho_s\) is parameterized by a multi-layer perceptron (MLP). After obtaining feature-channel contributions, HGNAN-edge combines them using a learnable global feature weight $\mathbf{w}\in \mathbb{R}^{p_e}$ and a scale weight \(\boldsymbol{\beta}\in\mathbb{R}^{s_{\max}}\) over adjacency orders, analogous to that in HGNAN-node:
\begin{equation}
\begin{aligned}
    z_{e_i}^{(s)}
    &=
    \sum_{k=1}^{p_e}
    w_k [\mathbf{g}_i^{(s)}]_k,
    &
    z_{e_i}
    &=
    \sum_{s=1}^{s_{\max}}
    \beta_s z_{e_i}^{(s)} .
\end{aligned}
\label{eq:hgnan_edge_gate}
\end{equation}

Equivalently, HGNAN-edge can be written as a decomposition into a feature response and a distance-aware structural influence term.
Define the feature-wise response \(\phi(e_j)\) and structural influence through \(s\)-intersection graph \(\kappa^{(s)}(e_i,e_j)\), and we have
\begin{align}
    \phi(e_j)
    &=
    \sum_{k=1}^{p}
    w_k f_k([\mathbf{y}_j]_k),\\
    \quad
    \kappa^{(s)}(e_i,e_j)
    &=
    \frac{1}{\# d^{(s)}(e_i,e_j)}
    \rho_s\!\left(r_{ij}^{(s)}\right)\\
    z_{e_i}^{(s)}
    &=
    \sum_{j:\, d^{(s)}(e_i,e_j)<\infty}
    \kappa_s(e_i,e_j)\phi(e_j).
\end{align}
This decomposition provides direct feature-level and distance-level interpretations for hyperedge prediction.

For multiclass prediction, the same decomposition is applied component-wise using vector-valued shape functions. Overall, HGNAN-edge extends additive modeling to the hyperedge domain by combining explicit hyperedge features with global, distance-aware structural effects, making each prediction directly traceable to both the attributes of the candidate hyperedge and the higher-order relations that influence it.

\subsection{Comparison Between HGNAN-node and HGNAN-edge}
HGNAN-node and HGNAN-edge share the same additive modeling principle, decomposing predictions into feature-wise shape functions and hypergraph structural influence, but differ in the structural domain where aggregation is performed. HGNAN-node aggregates over node neighborhoods induced by \(s\)-adjacency and is designed for node-level prediction, where local homophilic neighborhoods often contain task-relevant information; its learned neighbor weights further provide instance-specific neighbor-level attribution. In contrast, HGNAN-edge operates on the hyperedge domain induced by the \(s\)-intersection graph and targets hyperedge prediction settings, such as metabolic reaction prediction, where predictive evidence may extend beyond immediate local neighborhoods. It therefore uses global distance-aware aggregation to model how hyperedges at different structural distances contribute to a target prediction. These design choices are modular rather than intrinsic to the prediction level: in principle, local or global structural mechanisms can be paired with either node- or hyperedge-level tasks according to the structural characteristics of the application.

\subsection{Training Pipeline}
HGNAN adopts a unified supervised learning pipeline for both node- and hyperedge-level tasks. Given node features $\mathbf{X} \in \mathbb{R}^{n \times p}$ and incidence matrix $\mathbf{H}$, we first construct the node $s$-adjacency and the hyperedge $s$-intersection graph. For node prediction, HGNAN-node directly uses $(\mathbf{X}, \mathbf{H})$ and produces node embeddings $\{\mathbf{h}_i\}_{i=1}^n$ through feature-wise shape functions and neighborhood aggregation. These additive representations are passed through a linear prediction layer, preserving the additive structure of the final logits and optimized using cross-entropy or binary cross-entropy based on the number of classes, following standard settings used in canonical HGNNs. For hyperedge prediction, we compute hyperedge features $\mathbf{Y} = \{\Gamma(\{\mathbf{x}_l : v_l \in e_i\})\}_{i=1}^m$, and HGNAN-edge applies the same additive mechanism over the $s$-intersection graph to generate hyperedge embeddings $\{\mathbf{g}_e^{(s)}\}_{e=1}^m$, which are passed to a prediction head and trained with binary cross-entropy for hyperedge prediction. Importantly, all structural distances are computed solely from the original hypergraph and remain independent of the negative-sampling procedure.

\section{Experiments}\label{experiments}

We evaluate HGNAN from six complementary perspectives.
First, we test whether HGNAN-node maintains competitive predictive performance on standard hypergraph node-level prediction benchmarks, including both classical and recent HGNN baselines. Second, we evaluate HGNAN-edge on missing reaction recovery in genome-scale metabolic networks, a higher-order biochemical networked system where reactions are naturally modeled as hyperedges. Third, we conduct ablation studies to examine the contributions of feature-wise functions, hypergraph aggregation, and additive decomposition. Fourth, we examine whether HGNAN can recover known feature and structural mechanisms in controlled synthetic settings. Fifth, we evaluate whether the learned feature and structural functions transfer across related metabolic hypergraph systems. Finally, we analyze the transparent decompositions produced by HGNAN-node and HGNAN-edge through case studies on animal classification and metabolic reaction recovery. The code and data associated with HGNAN are available at https://github.com/ShihanF0/HGNAN.

\subsection{Node-Level Prediction on Hypergraph Benchmarks}

We first evaluate HGNAN on standard hypergraph node-level prediction benchmarks. The goal of this experiment is not to claim that an interpretable model is universally more accurate than all black-box HGNNs, but to verify that the proposed additive and interpretable structure does not lead to a substantial loss in predictive performance. We therefore compare HGNAN-node with both classical and recent hypergraph neural network baselines.

\subsubsection{Datasets and Baselines}
We evaluate on six node classification datasets, including four commonly used hypergraph benchmarks, Zoo, Mushroom, NTU2012, and Cora, and two heterophilic node classification datasets, Pokec and Actor. Dataset details are provided in Appendix~\ref{dataset_info}. We compare HGNAN with an MLP baseline, which does not use hypergraph structure, as well as representative HGNN baselines including HGNN~\cite{10.1609/aaai.v33i01.33013558}, HyperGCN~\cite{NEURIPS2019_1efa39bc}, AllDeepSets~\cite{Chien2021YouAA}, AllSetTransformer~\cite{Chien2021YouAA}, UniGCNII~\cite{ijcai2021p353}, ED-HNN~\cite{wang2022equivariant}, ZEN~\cite{bae2026parameter}, GeDi-HNN~\cite{Fiorini2024LetTB}, and KHGNN~\cite{10.1609/aaai.v39i20.35472}. These baselines cover both classical and recent developments in hypergraph learning. All methods are evaluated under the same data splits and evaluation protocol. Implementation details and hyperparameter search spaces are provided in Appendix~\ref{training_info}.

\subsubsection{Results}
Table~\ref{tab:node_results} reports test accuracy averaged over ten random splits. HGNAN-node attains the best average rank ($3.2$) among all compared methods and the best mean accuracy on NTU2012 and Mushroom. Several HGNN baselines are stronger on individual homophilic benchmarks, most notably ED-HNN on Cora and ZEN on Zoo, but HGNAN-node reaches this overall level of performance while exposing intrinsic feature-level and neighbor-level decompositions of its predictions. On the two heterophilic datasets, Pokec and Actor, the structure-agnostic MLP outperforms every hypergraph learning model. To test whether this phenomenon reflects the models or the data, we augment the same MLP with the mean features of each node's one-hop and two-hop hypergraph neighbors, keeping the remaining unchanged. The augmentation lowers accuracy on both datasets by 1.5\% and 2.2\%, respectively, indicating that the hypergraph structure carries no usable label information on these benchmarks. This finding is consistent with previous literature on both graph and hypergraph \cite{zhu2020beyond,Li_Gu_Wang_Fang_Bai_Zhuang_Liò_2025}. The best hypergraph models, HGNAN-node included, trail the MLP by a margin of the same size ($1.0$ and $1.4$ accuracy points for HGNAN-node on Pokec and Actor), which is also consistent with the negligible effect of removing aggregation on Pokec in Section~\ref{sec:ablation}.

\begin{table*}[t]
\centering
\caption{Comparison of test accuracy (mean $\pm$ standard deviation) between HGNAN-node and baselines across node classification datasets. Bold indicates the highest test accuracy. HGNAN achieves best average rank among all compared methods.}
\label{tab:node_results}
\renewcommand{\arraystretch}{1.2}
\resizebox{\textwidth}{!}{
\begin{tabular}{c c c c c c c c}
\toprule
\textbf{Method} & \textbf{Zoo} & \textbf{Mushroom} & \textbf{NTU2012} & \textbf{Cora} & \textbf{Pokec} & \textbf{Actor} & \textbf{Avg. Rank} \\
\midrule
MLP & 0.935 $\pm$ 0.035 & \textbf{1.000 $\pm$ 0.000} & 0.883 $\pm$ 0.012 & 0.744 $\pm$ 0.019 & \textbf{0.597 $\pm$ 0.007} & \textbf{0.868 $\pm$ 0.005} & 4.8 \\
\midrule
AllDeepSets & 0.946 $\pm$ 0.031 & 0.999 $\pm$ 0.001 & 0.876 $\pm$ 0.011 & 0.767 $\pm$ 0.016 & 0.587 $\pm$ 0.006 & 0.847 $\pm$ 0.007 & 6.5 \\
AllSetTransformer & 0.969 $\pm$ 0.041 & 0.999 $\pm$ 0.001 & 0.884 $\pm$ 0.010 & 0.792 $\pm$ 0.020 & 0.587 $\pm$ 0.007 & 0.855 $\pm$ 0.005 & 3.4 \\
HGNN & 0.950 $\pm$ 0.042 & 0.998 $\pm$ 0.001 & 0.872 $\pm$ 0.017 & 0.784 $\pm$ 0.012 & 0.580 $\pm$ 0.006 & 0.768 $\pm$ 0.005 & 7.5 \\
HyperGCN & 0.423 $\pm$ 0.000 & 0.482 $\pm$ 0.000 & 0.765 $\pm$ 0.031 & 0.781 $\pm$ 0.022 & 0.581 $\pm$ 0.008 & 0.623 $\pm$ 0.000 & 9.7 \\
UniGCNII & 0.958 $\pm$ 0.047 & 0.999 $\pm$ 0.001 & 0.891 $\pm$ 0.011 & 0.787 $\pm$ 0.014 & 0.587 $\pm$ 0.008 & 0.829 $\pm$ 0.006 & 4.3 \\
\midrule
ED-HNN & 0.938 $\pm$ 0.043 & 0.999 $\pm$ 0.001 & 0.886 $\pm$ 0.014 & \textbf{0.807 $\pm$ 0.016} & 0.585 $\pm$ 0.004 & 0.840 $\pm$ 0.006 & 5.2 \\
ZEN & \textbf{0.979 $\pm$ 0.021} & 0.856 $\pm$ 0.007 & 0.863 $\pm$ 0.011 & 0.728 $\pm$ 0.021 & 0.541 $\pm$ 0.011 & 0.593 $\pm$ 0.009 & 8.9 \\
GeDi-HNN & 0.954 $\pm$ 0.039 & \textbf{1.000 $\pm$ 0.000} & 0.885 $\pm$ 0.017 & 0.781 $\pm$ 0.019 & 0.593 $\pm$ 0.005 & 0.859 $\pm$ 0.004 & 3.9 \\
KHGNN & 0.892 $\pm$ 0.042 & 0.995 $\pm$ 0.006 & 0.863 $\pm$ 0.015 & 0.790 $\pm$ 0.016 & 0.579 $\pm$ 0.006 & 0.659 $\pm$ 0.006 & 8.4 \\
\midrule
HGNAN (ours) & 0.958 $\pm$ 0.032 & \textbf{1.000 $\pm$ 0.000} & \textbf{0.896 $\pm$ 0.011} & 0.787 $\pm$ 0.019 & 0.587 $\pm$ 0.008 & 0.854 $\pm$ 0.004 & \textbf{3.2} \\
\bottomrule
\end{tabular}
}
\end{table*}

\subsection{Hyperedge-Level Prediction on Metabolic Networks}
\label{sec:metabolic_hyperedge_prediction}

\begin{table}[t]
\centering
\caption{%
Comparison of hyperedge prediction AUROC (mean $\pm$ standard deviation)
across four BiGG GEM datasets: iAF1260b, iJR904, iSB619, and iYO844.
Bold highlights the best result for each dataset.
HGNAN outperforms all baseline models across all datasets.
}
\renewcommand{\arraystretch}{1.4}
\setlength{\tabcolsep}{3pt}
\resizebox{\linewidth}{!}{%
\begin{tabular}{ccccc}
\toprule
\textbf{Method}
& \textbf{iAF1260b}
& \textbf{iJR904}
& \textbf{iSB619}
& \textbf{iYO844} \\
\midrule
CHESHIRE
& $0.814 \pm 0.011$
& $0.836 \pm 0.017$
& $0.845 \pm 0.023$
& $0.883 \pm 0.013$ \\
NHP
& $0.723 \pm 0.018$
& $0.729 \pm 0.022$
& $0.720 \pm 0.021$
& $0.797 \pm 0.015$ \\
HyperSAGNN
& $0.759 \pm 0.029$
& $0.784 \pm 0.011$
& $0.777 \pm 0.020$
& $0.830 \pm 0.028$ \\
\midrule
HGNAN (ours)
& $\mathbf{0.914 \pm 0.010}$
& $\mathbf{0.908 \pm 0.021}$
& $\mathbf{0.904 \pm 0.010}$
& $\mathbf{0.908 \pm 0.012}$ \\
\bottomrule
\end{tabular}%
}
\label{tab:edge_result}
\end{table}

We next evaluate HGNAN on missing reaction recovery in genome-scale metabolic models (GEMs). GEMs are widely used to model cellular metabolism and predict physiological states~\cite{Fang2020}. However, even well-curated GEMs often contain knowledge gaps, including missing biochemical reactions. Predicting missing reactions can be formulated as a hyperedge prediction problem: metabolites are represented as nodes, and biochemical reactions are represented as hyperedges connecting the participating metabolites~\cite{chen2023teasing,Chen2023Survey}. This setting provides real biochemical networked systems in which both predictive performance and mechanism-level interpretability are important.

\subsubsection{Datasets and Baselines}
We evaluate on four GEMs from the BiGG database~\cite{King2016}: iAF1260b, iJR904, iSB619, and iYO844. Detailed dataset information is provided in Appendix~\ref{dataset_info}.
We compare HGNAN-edge with representative hyperedge prediction baselines, including HyperSAGNN~\cite{zhang2020hypersagnn}, NHP~\cite{NHP}, and CHESHIRE~\cite{chen2023teasing}, which have been used for missing reaction recovery in metabolic networks.
Because BiGG models do not provide inherent metabolite-level features, we construct chemically interpretable metabolite features using MACCS Keys~\cite{RDKit}. Each MACCS key corresponds to a predefined molecular substructure or functional group, so a metabolite \(v\) is represented by a binary fingerprint \(\mathbf{x}_v\in\{0,1\}^{167}\), where \([\mathbf{x}_v]_k=1\) indicates the presence of the \(k^{\mathrm{th}}\) functional group. For a reaction hyperedge \(e_i\), we then construct a hyperedge-level feature vector by comparing the functional groups before and after the reaction. Specifically, we define
\begin{equation} \mathbf{y}_i = \sum_{v\in\mathcal{P}(e_i)} \mathbf{x}_v - \sum_{v\in\mathcal{R}(e_i)} \mathbf{x}_v, \label{eq:reaction_feature} \end{equation}
where \(\mathcal{P}(e_i)\) and \(\mathcal{R}(e_i)\) denote the sets of product and reactant metabolites in reaction \(e_i\) respectively, so that \([\mathbf{y}_i]_k\) represents the net change in the count of the \(k^{\mathrm{th}}\) functional group induced by reaction \(e_i\). Positive values indicate functional groups generated by the reaction, whereas negative values indicate functional groups consumed by the reaction. This construction turns each reaction into a hyperedge with interpretable features, allowing HGNAN-edge to learn directly from biologically meaningful reaction attributes rather than relying only on learned node embeddings after message passing. Details of metabolite feature generation and missing-feature handling are provided in Appendix~\ref{dataset_info}.
Following standard hyperedge prediction protocols, we generate negative hyperedges by replacing a subset of metabolites in observed reactions with randomly sampled metabolites from the metabolite pool. We use the same dataset splits and evaluation protocol across all methods. Additional metrics, including accuracy, AUPRC, and F1 score, as well as hyperparameter search details, are provided in Appendix~\ref{training_info}.

\subsubsection{Results}
Table~\ref{tab:edge_result} reports hyperedge prediction accuracy on the four GEM datasets. HGNAN achieves the highest mean AUROC on all four datasets, with particularly large improvements on iAF1260b and iJR904. These results indicate that directly modeling hyperedge-level features and distance-aware structural influence is effective for missing higher-order interaction recovery in metabolic networks. Note that the baselines here use only the incidence structure, as in their original implementations. Appendix~\ref{app:edge_maccs} reports a comparison in which every baseline additionally receives the same node-level MACCS fingerprints. Beyond predictive accuracy, HGNAN provides direct feature-level and distance-level decompositions of each prediction.
This is important for metabolic network analysis, where one may want to understand not only whether a candidate reaction is plausible, but also which chemical functional groups and structural relations support the prediction. We analyze these decompositions in Section~\ref{sec:transparent_decomposition} through a case study on iAF1260b, showing how HGNAN-edge combines functional-group effects with distance-aware structural influence to explain missing reaction predictions.

\subsection{Ablation Study}
\label{sec:ablation}

We conduct ablation studies to evaluate three key components of HGNAN: hypergraph aggregation, feature-wise additive decomposition, and input features.
Experiments are conducted on four representative node classification datasets (Zoo, NTU2012, Cora, and Pokec) and two representative hyperedge prediction datasets (iJR904 and iSB619).
We consider the following variants:
\begin{itemize}
    \item \textbf{w/o Aggregation (NAM)}: It removes hypergraph aggregation, reducing HGNAN to a NAM-style model that predicts from feature-wise shape functions only.
    \item \textbf{w/o Additive Structure}: It replaces feature-wise shape functions with a single MLP over all input features, removing the additive decomposition while preserving the corresponding hypergraph aggregation mechanism.
    \item \textbf{w/o Node Features}: It replaces all feature inputs with a shared learnable constant vector, so predictions rely only on hypergraph structural information.
\end{itemize}
Results are reported as mean \(\pm\) standard deviation over 10 runs in Table~\ref{tab:ablation_all}.

\begin{table*}[t]
\centering
\caption{
Ablation study across node classification and hyperedge prediction datasets.
Results are AUROC/AUPRC (\%) reported as mean \(\pm\) standard deviation over 10 runs.
\textbf{Bold} indicates the best result in each column.
}
\label{tab:ablation_all}
\small
\setlength{\tabcolsep}{3pt}
\renewcommand{\arraystretch}{1.10}
\resizebox{\textwidth}{!}{
\begin{tabular}{lcccccccccccc}
\toprule
& \multicolumn{2}{c}{\textbf{Zoo}} 
& \multicolumn{2}{c}{\textbf{NTU2012}} 
& \multicolumn{2}{c}{\textbf{Cora}} 
& \multicolumn{2}{c}{\textbf{Pokec}} 
& \multicolumn{2}{c}{\textbf{iJR904}} 
& \multicolumn{2}{c}{\textbf{iSB619}} \\
\cmidrule(lr){2-3}
\cmidrule(lr){4-5}
\cmidrule(lr){6-7}
\cmidrule(lr){8-9}
\cmidrule(lr){10-11}
\cmidrule(lr){12-13}
\textbf{Model} 
& \textbf{AUROC} & \textbf{AUPRC} 
& \textbf{AUROC} & \textbf{AUPRC} 
& \textbf{AUROC} & \textbf{AUPRC} 
& \textbf{AUROC} & \textbf{AUPRC} 
& \textbf{AUROC} & \textbf{AUPRC} 
& \textbf{AUROC} & \textbf{AUPRC} \\
\midrule

HGNAN (full)
    & \(\mathbf{98.40}_{\pm 1.90}\)& \(\mathbf{94.11}_{\pm 5.19}\)& \(\mathbf{99.56}_{\pm 0.29}\)
    & \(\mathbf{93.18}_{\pm 1.06}\)
    & \(\mathbf{96.06}_{\pm 0.71}\)
    & \(\mathbf{85.30}_{\pm 2.11}\)
    & \(64.38_{\pm 0.69}\)
    & \(66.86_{\pm 0.76}\)
    & \(\mathbf{90.81}_{\pm 2.07}\)
    & \(\mathbf{87.54}_{\pm 3.12}\)
    & \(\mathbf{90.44}_{\pm 1.03}\)
    & \(\mathbf{87.39}_{\pm 1.95}\) \\

w/o Aggregation (NAM)
    & \(97.95_{\pm 1.90}\)& \(93.56_{\pm 5.16}\)& \(98.51_{\pm 0.31}\)& \(92.23_{\pm 1.02}\)& \(93.84_{\pm 0.86}\)
    & \(77.74_{\pm 2.37}\)
    & \(63.84_{\pm 0.66}\)& \(66.63_{\pm 0.73}\)& \(88.72_{\pm 0.80}\)& \(86.48_{\pm 1.18}\)& \(87.16_{\pm 1.60}\)
    & \(84.21_{\pm 2.73}\) \\

w/o Additive Structure
    & \(98.32_{\pm 1.74}\)
    & \(93.17_{\pm 6.23}\)
    & \(99.33_{\pm 0.16}\)
    & \(87.96_{\pm 0.89}\)
    & \(95.19_{\pm 0.71}\)
    & \(83.27_{\pm 2.06}\)
    & \(\mathbf{64.64}_{\pm 0.68}\)
    & \(\mathbf{67.17}_{\pm 0.77}\)
    & \(89.55_{\pm 1.98}\)
    & \(85.24_{\pm 2.98}\)
    & \(82.16_{\pm 11.35}\)
    & \(79.83_{\pm 11.58}\) \\

w/o Node Features
    & \(52.21_{\pm 5.69}\)
    & \(17.54_{\pm 3.01}\)
    & \(54.94_{\pm 6.07}\)
    & \(4.15_{\pm 0.88}\)
    & \(51.71_{\pm 2.06}\)
    & \(16.70_{\pm 1.30}\)
    & \(51.17_{\pm 1.01}\)
    & \(51.45_{\pm 0.67}\)
    & \(61.20_{\pm 1.66}\)
    & \(60.12_{\pm 1.75}\)
    & \(62.24_{\pm 3.34}\)
    & \(60.72_{\pm 4.08}\) \\

\bottomrule
\end{tabular}
}
\end{table*}

On node classification datasets, removing node features leads to AUROC values close to random and sharply reduced AUPRC, indicating that these benchmarks are primarily driven by node attributes. Removing aggregation has a smaller and dataset-dependent effect on node classification tasks. It gives nearly identical results on Zoo and Pokec, slightly decreases AUROC and AUPRC on NTU2012, and causes a clear drop on Cora ($-2.22$ AUROC and $-7.56$ AUPRC). In contrast, on hyperedge prediction datasets, removing aggregation leads to a substantial drop, showing that distance-aware structural aggregation provides useful information beyond chemical features.
The non-additive variant also underperforms the full model on most datasets, suggesting that the additive decomposition is useful not only for interpretability but also for predictive performance or stability.
Overall, the ablation study indicates that HGNAN benefits from combining feature-wise nonlinear responses with hypergraph structural influence and that these components play complementary roles depending on the task.

\subsection{Feature--Structure Recovery}
\label{sec:synthetic_recovery}

We next evaluate whether HGNAN-edge can recover known feature and structural mechanisms under a controlled data-generating process.
This experiment tests whether the learned shape functions and distance-response function recover the underlying mechanism, rather than merely yielding visually plausible decompositions. We construct a synthetic hyperedge-level classification task whose data-generating process follows the same additive decomposition of feature responses and structural influence as HGNAN-edge. The synthetic hypergraph contains two sparse hyperedge clusters connected by a chain, creating both local and long-range hyperedge distances. Each node is assigned a two-dimensional feature vector sampled from a uniform distribution, and each hyperedge feature is computed from the features of its member nodes.
The ground-truth feature mechanisms $f_1, f_2$ and ground-truth structural mechanism $\rho$ on $1$-intersection graph are defined as
\(
f_1(x_1)=-x_1,\
f_2(x_2)=\sin(\pi x_2),\
\rho(d)=\exp(-d/2)
\) respectively,
with feature weights \(w_1=2\) and \(w_2=1\). Hyperedge logits are generated by combining these feature and structural components using the HGNAN-edge aggregation form, followed by standardization and Gaussian noise.

\begin{figure}[t]
    \centering
    \includegraphics[width=\linewidth]{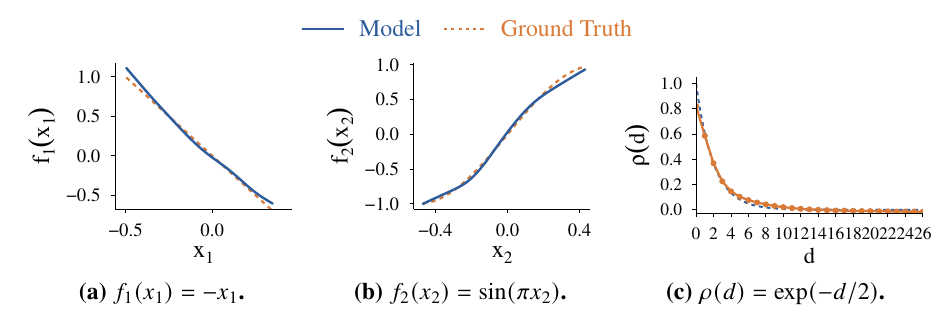}
    \caption{Controlled recovery of ground-truth feature and structural mechanisms on the synthetic hyperedge prediction task.
    HGNAN-edge closely recovers the two feature-wise shape functions and the distance-based structural influence function.
    This demonstrates that the learned decompositions can recover the underlying data-generating mechanism in a controlled setting.}
    \label{fig:synthetic_figure}
\end{figure}

To compare learned mechanisms with ground truth, we align the learned functions to the ground-truth functions using an affine transformation to account for scale and shift differences, and report mean squared error \(MSE\), \(R^2\), and Pearson correlation. Fig.~\ref{fig:synthetic_figure} shows the original data-generating mechanism and the mechanism recovered by HGNAN-edge. HGNAN-edge accurately recovers both feature functions and the distance function: the aligned curves achieve MSE values between \(5\times 10^{-4}\) and \(1.4\times 10^{-3}\), with \(R^2\) and Pearson correlations close to \(0.99\) for all recovered mechanisms.
These results show that the decomposable architecture of HGNAN-edge can recover feature-wise contributions and distance-based structural influence when the underlying data-generating process is known.

\subsection{Cross-Network Transfer Across Metabolic Hypergraphs}
\label{sec:cross_bigg_transfer}

We further evaluate whether the learned feature and structural functions of HGNAN-edge transfer across related metabolic hypergraph systems. This experiment is designed to test whether HGNAN-edge learns reusable reaction-level mechanisms, rather than patterns that are specific to a single GEM dataset.
We perform cross-dataset validation among the four BiGG GEM datasets under zero-shot cross-dataset transfer setting.
For each source-target pair, we first train a model on the source GEM, freeze all learned parameters and directly evaluate the model on the target GEM. The model is then evaluated on the target test set using AUROC. For HGNAN-edge, this transfer is natural because the MACCS-based feature functions and distance functions have consistent input dimensions across GEMs. To make baseline transfer feasible, we pad the incidence matrices to a shared metabolite universe before transfer. Table~\ref{tab:crossdataset} in Appendix~\ref{app:transfer} shows that HGNAN-edge transfers effectively across metabolic hypergraphs.
Across the 12 source-target pairs, HGNAN-edge achieves the best on 8 pairs. For example, when trained on iAF1260b and transferred to iJR904, HGNAN-edge achieves an AUROC of \(0.899\).
These results suggest that the feature-wise shape functions and distance-aware structural functions learned by HGNAN-edge capture transferable biochemical regularities across related GEMs.

\subsection{Quantitative and Qualitative Interpretability Evaluation}
\label{sec:transparent_decomposition}

We now examine how HGNAN decomposes its predictions into interpretable feature and structural contributions. HGNAN can be interpreted through two complementary components. First, the feature-wise shape functions \(w_kf_k(\cdot)\) characterize the nonlinear effects of individual input features. Second, the structural components determine how these feature effects are modulated and aggregated according to the underlying hypergraph structure. For HGNAN-node, the structural component is given by sparse neighbor weights \(a_{ij}^{(s)}\), which act as adaptive spatial aggregation coefficients over local hypergraph neighborhoods. For HGNAN-edge, the structural component is given by distance-aware functions \(\rho_s(\cdot)\), which act as distance-dependent structural influence functions over the \(s\)-intersection graph. We illustrate these decompositions through case studies on node classification and metabolic reaction recovery.

\subsubsection{Node-Level Interpretability}

\begin{figure}[t]
    \centering
    \includegraphics[width=\linewidth]{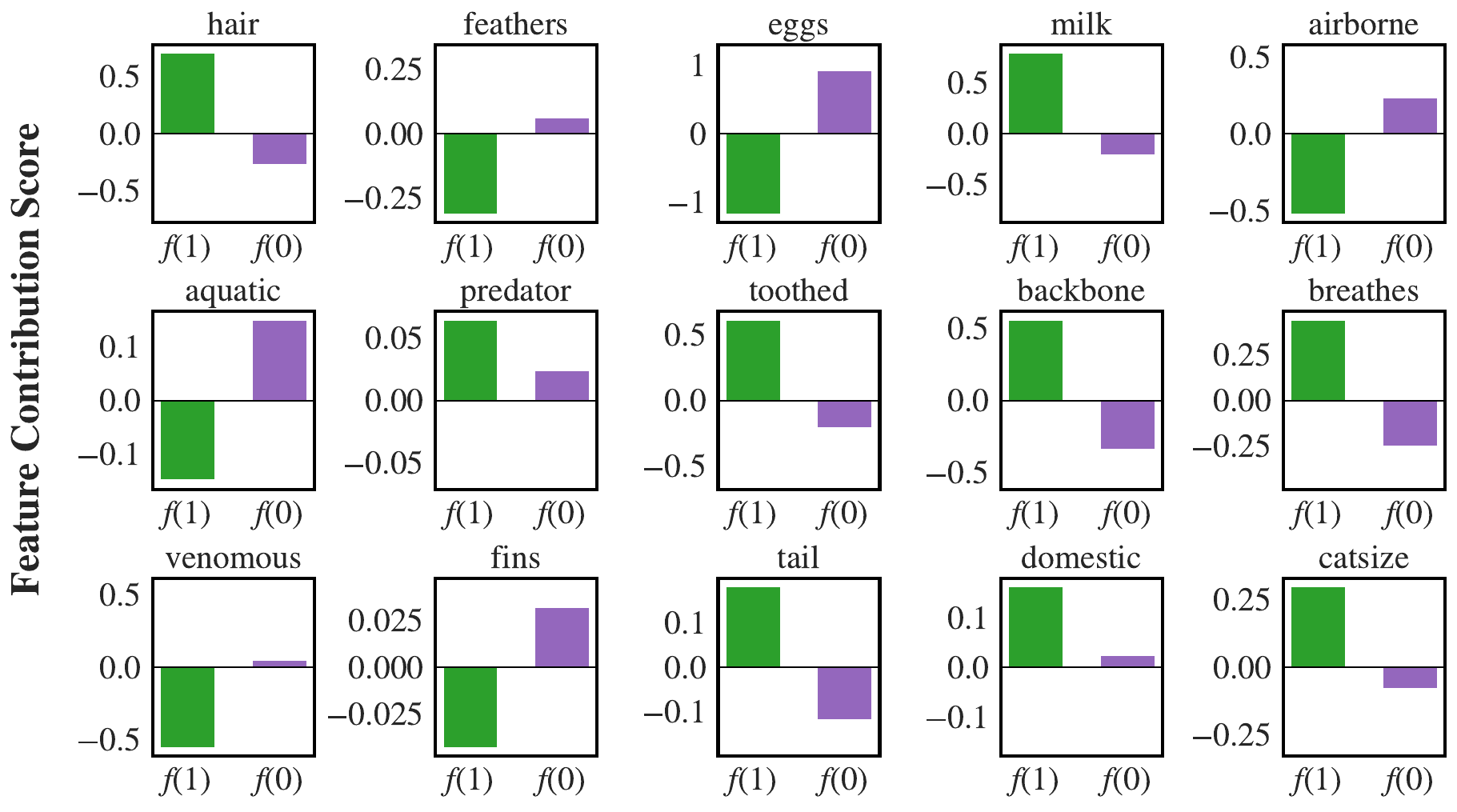}
    \caption{
    Feature contribution scores for predicting the ``Mammal'' class in the Zoo dataset. Each bar plot shows the learned response \(w_k f_k(x_k)\) for a binary feature, where \(x=1\) indicates feature presence and \(x=0\) indicates feature absence. The feature ``milk'' strongly supports mammal classification, whereas ``feathers'' suppresses it, consistent with biological knowledge of mammalian traits.
    }
    \label{fig:zoo_visualize}
\end{figure}

\begin{figure}[t]
    \centering
    \includegraphics[width=1\linewidth]{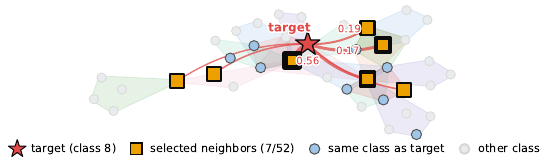}
    \caption{
    Local structural decomposition for node 232 in the NTU2012 dataset. The red star denotes the target node, and squares with black outlines are the neighbors that receive a non-zero aggregation weight \(\sum_s \beta_s a^{(s)}_{ij}\), with thicker outlines and thicker arcs indicating larger weights. The three largest weights are annotated. The blue circle are neighbours that share the same class as the target but not receiving weights. White circles are nodes belongs to other classes and not selected. Shaded regions are hyperedges. HGNAN-node assigns a non-zero weight to only \(7\) of the \(52\) nodes in the neighborhood of the target. It shows that HGNAN-node can suppress misleading local evidence and concentrate aggregation on structurally and semantically consistent neighbors.
    }
    \label{fig:node_interpretation_232}
\end{figure}

We first illustrate feature-wise responses on the Zoo dataset by analyzing predictions for the ``Mammal'' class. For each binary feature \(k\), we plot the learned responses \(w_k f_k(1)\) and \(w_k f_k(0)\), corresponding to feature presence and absence, respectively. The sign of \(w_k f_k(x_k)\) indicates whether the feature value supports or suppresses the mammal prediction, while the magnitude reflects the strength of the response. Fig.~\ref{fig:zoo_visualize} shows that ``milk'' has the largest positive response when present, while ``feathers'' has a negative response for the mammal score. These patterns align with biological knowledge that producing milk is a defining mammalian trait, whereas feathers are associated with birds rather than mammals. This example shows that HGNAN-node preserves the pointwise feature-response interpretability of additive models while extending them to hypergraph data. 

We next examine the local structural decomposition on NTU2012. Since HGNAN-node uses sparse adaptive weights \(a_{ij}^{(s)}\) for neighborhood aggregation, it can identify which neighbors form the effective local support for a target node prediction. Fig.~\ref{fig:node_interpretation_232} visualizes the learned aggregation support for a representative target node. The red star denotes the target node, and highlighted neighbors correspond to nodes receiving non-negligible aggregation weights. In this example, HGNAN-node assigns a non-zero weight to only \(7\) of the \(52\) nodes in the neighborhood of the target, yielding a sparse local aggregation pattern. Across the ten runs the number of selected neighbors ranges from \(5\) to \(13\) with a median of \(7\). All seven selected neighbors carry the target's class, although only \(15\) of the \(52\) candidate neighbors do, and the target is classified correctly in all ten runs. This suggests that HGNAN-node can suppress misleading local evidence and concentrate aggregation on structurally and semantically consistent neighbors. Together, the Zoo and NTU2012 examples show that HGNAN-node exposes both pointwise feature responses and adaptive structural aggregation patterns, making its node-level predictions directly inspectable.

\subsubsection{Hyperedge-Level Interpretability.}

\begin{figure*}[t]
    \centering
    \includegraphics[width=\linewidth]{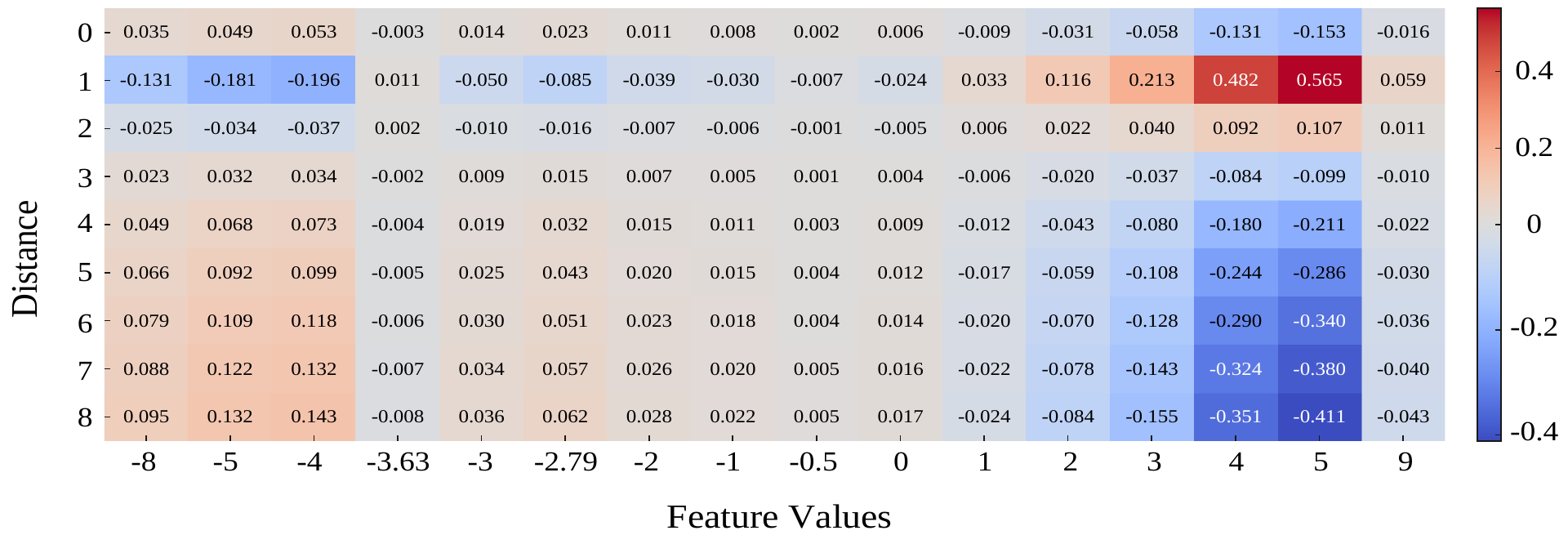}
    \caption{Heatmap for Carbonyl group. The horizontal axis denotes the hyperedge feature value \([\mathbf{y}_j]_{\mathrm{C=O}}\), corresponding to the net change in Carbonyl-group count. The vertical axis denotes hyperedge distance \(d^{(1)}(e_i,e_j)\). Each cell shows the contribution induced by a specific chemical input value and structural-distance combination.}
    \label{fig:c-o heatmap}
\end{figure*}
\begin{figure}[t]
    \centering
\includegraphics[width=\linewidth]{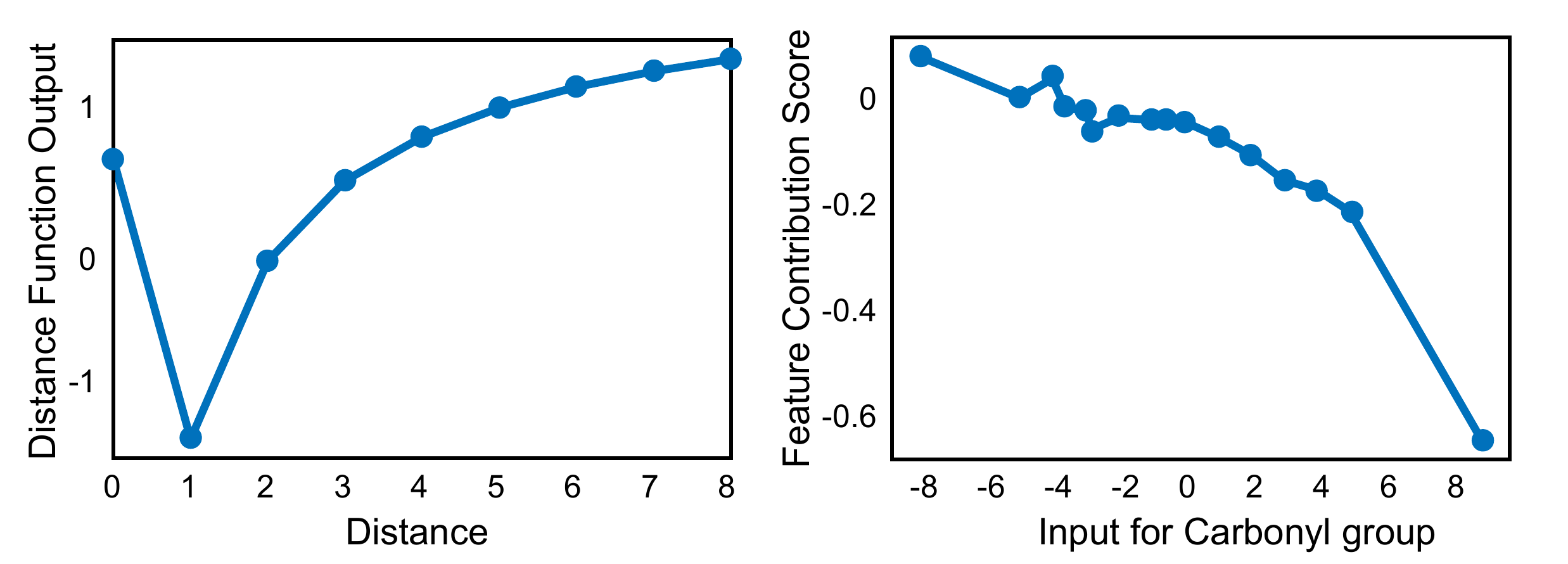}
    \caption{
    Hyperedge-level contribution decomposition on iAF1260b. Left: learned structural contribution \(\rho_1(1/(1+d))\) as a function of hyperedge distance \(d\). Right: feature-wise contribution \(w_k f_k(\cdot)\) for the Carbonyl group \((\mathrm{C{=}O})\). Linear interpolation is used only for visualization.}
    \label{fig:dis and feature for iaf1260b}
\end{figure}

HGNAN-edge provides hyperedge-level interpretations by decomposing each prediction into feature-wise nonlinear responses and distance-aware structural responses. We illustrate this on the iAF1260b dataset from the BiGG database. As described in Section~\ref{sec:metabolic_hyperedge_prediction}, each metabolite is represented by chemically interpretable functional-group features, and each reaction hyperedge is represented by the net change in functional-group counts before versus after the reaction. For iAF1260b and \(s_{\max}=1\), we visualize the learned structural response \(\rho_1(\cdot)\). For interpretability, we plot the effective distance-domain response \(\rho_1(1/(1+d))\) as a function of hyperedge distance \(d=d_1(e_i,e_j)\). We also visualize the feature-wise nonlinear response \(w_k f_k(\cdot)\) for the Carbonyl group \((\mathrm{C{=}O})\). 
Fig.~\ref{fig:dis and feature for iaf1260b} shows that the learned distance-domain response follows a V-shaped pattern. Hyperedges at distance \(0\), corresponding to the target reaction itself, receive a positive structural response. One-hop neighboring reactions receive a negative response, and this negative effect weakens as distance increases, eventually becoming mildly positive. This distance-response pattern is both empirically supported and biologically plausible. The positive response at distance \(0\) is consistent with the fact that the target reaction's own functional-group changes are directly informative for the reaction prediction. The negative response at distance \(1\) suggests that immediately adjacent reactions sharing metabolites may provide contrastive evidence, potentially reflecting competing branches or alternative transformations around shared precursors. At larger distances, the response becomes weakly positive. This agrees with the distance-based label-consistency analysis in Appendix~\ref{dataset_info}, where hyperedges at larger hyperedge distances in BiGG models tend to exhibit higher same-label ratios, although their coverage is smaller. Biologically, weak positive long-range influence indicates that reactions that are not locally adjacent still participate in coherent pathway-level or system-level metabolic organization, such as shared functional roles, biomass-related constraints, or broader stoichiometric compatibility. Thus, the learned distance response can be viewed as capturing both local contrast among nearby reactions and non-local regularities across the metabolic network. 

The Carbonyl response \(w_k f_k(\cdot)\) describes how changes in the Carbonyl-group input feature affect the prediction score before structural modulation. In this learned function, negative or small changes have limited effect, whereas positive increases in \(\mathrm{C{=}O}\) yield larger responses. The final contribution of a source hyperedge depends jointly on the transformed chemical inputs and the structural response determined by its distance from the target hyperedge. Fig.~\ref{fig:c-o heatmap} visualizes the joint feature-structure heatmap for the Carbonyl group. The same chemical feature can contribute differently depending on its structural location in the intersection graph. For example, the model assigns positive contributions to several combinations involving a loss of Carbonyl groups, while other feature-distance combinations produce different responses. The value of this decomposition is that it gives domain experts direct access to the learned mechanism: they can inspect the pointwise chemical response \(w_k f_k(\cdot)\), the distance-domain structural response \(\rho_s(\cdot)\), and their combined effect. If a learned response is inconsistent with biochemical knowledge, the corresponding component can be examined or adjusted without treating the entire predictor as a black box.

\subsubsection{Quantitative Faithfulness Evaluation}
We evaluate whether the components identified by HGNAN as important are faithful to the model's actual predictions. Following the comprehensiveness and sufficiency framework \cite{deyoung2020eraser}, \emph{comprehensiveness} measures the prediction degradation after removing the most important components, whereas \emph{sufficiency} measures the degradation when only those components are retained. We report COMP and SUFF, using the top 20\% important components. Higher COMP and lower SUFF indicate more faithful explanations. We compare feature attributions with Gradient$\times$Input (G$\times$I) \cite{ancona2017towards} and random attribution.

For node-level feature attribution, HGNAN shows particularly strong faithfulness on NTU2012, achieving average COMP of $0.502$, compared with $0.361$ for G$\times$I and $0.074$ for random attribution. Its SUFF is also substantially lower ($0.232$ versus $0.391$ and $0.766$, respectively). For structural attribution, removing the highest-weight neighbor on NTU2012 causes a $12.2$ times larger probability decrease than random removal, and the ratio remains $3.0$ times larger when the top ten neighbors are removed.

For HGNAN-edge, feature-level attributions are far more faithful than random attribution on all four networks, and compare favorably with G$\times$I in terms of sufficiency, although G$\times$I attains higher comprehensiveness on iJR904 and iYO844 (Appendix~\ref{app:faithfulness}). Structural faithfulness is substantially clearer because each source-hyperedge term $\kappa^{(s)}(e_i,e_j)\phi(e_j)$ is an exact additive contribution. On iAF1260b and iSB619, removing all non-self structural context reduces AUROC by $7.17$ and $6.26$ points, respectively. Removing only the ten largest-magnitude non-self contributions already reduces AUROC by $4.70$ and $5.04$ points, accounting for approximately $66\%$ and $81\%$ of the full contextual effect, while removing ten random source hyperedges changes AUROC by only $0.00$ and $0.02$ points. These results show that HGNAN's structural explanations identify a small set of components that account for most of the model's contextual dependence.

\subsection{Computation Complexity Analysis}
\label{app:time_complexity}
\subsubsection{Theoretical Analysis of Computational Cost}
We analyze the per-iteration cost of HGNAN. Let $N$ be the batch size, $p$ the input feature dimension, $d$ the output dimension, $h$ the hidden-layer width, $L$ the number of MLP layers, and $S$ the number of interaction scales. HGNAN consists of feature-wise shape networks $\{f_k\}_{k=1}^p$, a distance network $\rho$, and, for neighborhood-level aggregation, an attention MLP. Each shape network maps $\mathbb{R}\to\mathbb{R}^d$ and has
\[
\#\theta_f=
\begin{cases}
2d, & L=1,\\[2pt]
(1\cdot h+h)+(L-2)(h^2+h)+(hd+d), & L\ge 2.
\end{cases}
\]
Thus the feature-shape parameters are $\#\Theta_f=p\,\#\theta_f$. The distance network has the same order, multiplied by $S$, while the attention MLP contributes only $O(dh)$ additional parameters. Therefore, the dominant parameter cost is $\Theta(p\,\#\theta_f)$, which is linear in the input feature dimension $p$.

In each forward pass, the feature-wise MLPs are evaluated once per feature, yielding
\[
T_f=
\begin{cases}
\Theta(pNd), & L=1,\\[2pt]
\Theta\big(pN((L-2)h^2+hd)\big), & L\ge 2.
\end{cases}
\]
The aggregation cost depends on whether HGNAN-edge or HGNAN-node is used. For HGNAN-edge, overall aggregation applies $\rho$ to all $N^2$ pairwise distances at each interaction scale, resulting in
\[
T_{\mathrm{edge}}
=
\Theta \big(SN^2((L-2)h^2+h d_m+d)\big).
\]
The quadratic dependence on $N$ can be reduced in practice by low-rank approximations, such as Nystr\"{o}m approximation, to the $N\times N$ distance matrix. For HGNAN-node, attention is computed over $E$ neighbor edges followed by sparse aggregation, so the dominant cost is
\(
T_{\mathrm{node}}
=
\Theta(S E d H_a).
\)
\subsubsection{Practical Analysis of Computational Cost}

\begin{figure}[t]
    \centering
    \includegraphics[width=0.95\linewidth]{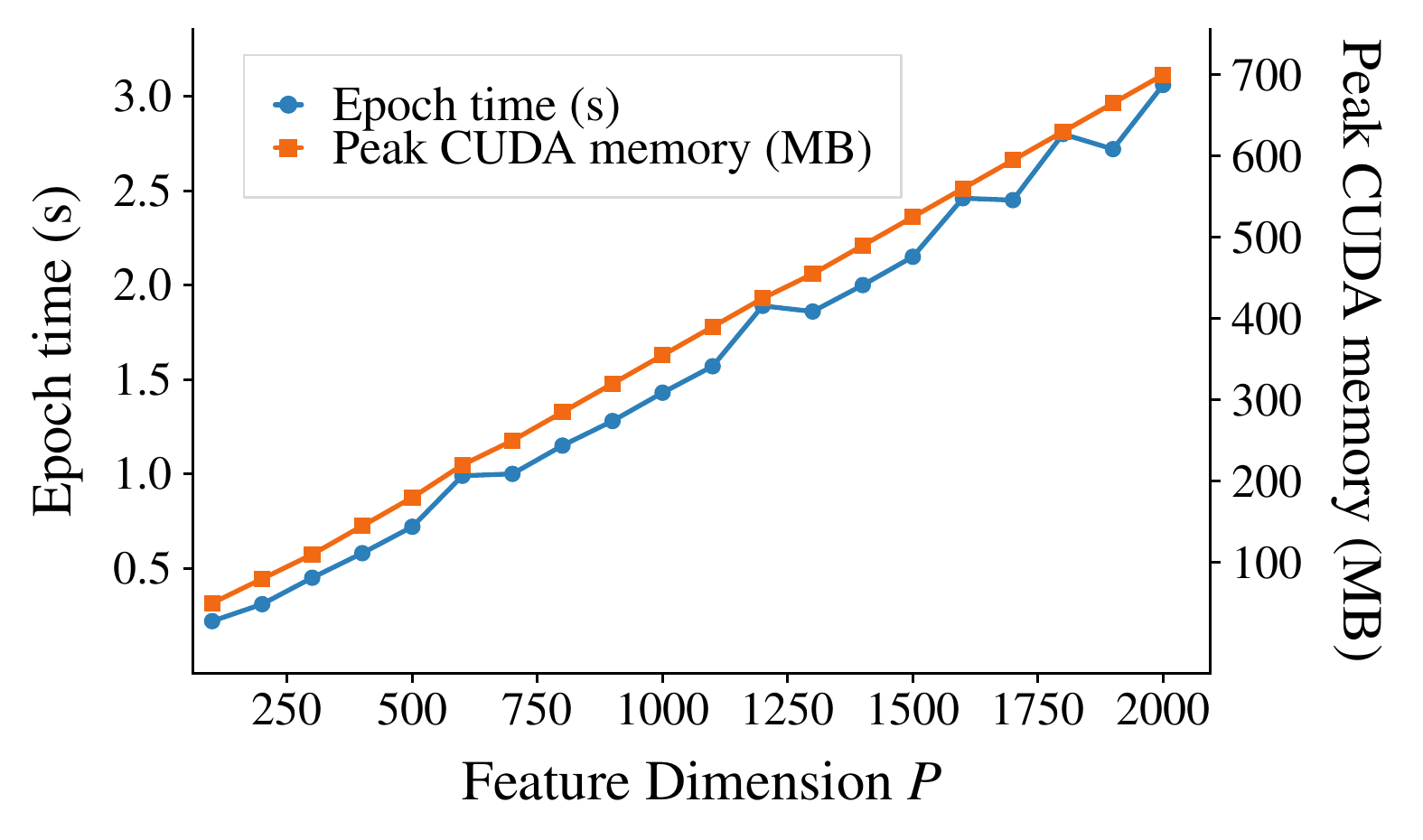}
    \caption{Empirical scaling of computational cost with feature dimension. We expand Zoo node features to dimension $P$ by concatenating repeated copies of the original features, and measure the training time per epoch and peak CUDA memory for one training epoch. Both metrics grow approximately linearly with $P$, supporting the theoretical linear-in-$P$ efficiency in practice.}
    \label{fig:Practical_Computational_Cost}
\end{figure}

We conduct an additional experiment to empirically verify that the computational cost of HGNAN scales linearly with the input feature dimension. Starting from the Zoo dataset, we construct a family of synthetic feature matrices by repeatedly concatenating copies of the original features until the feature dimension reaches $P$. We vary $P$ from 100 to 2000, train HGNAN for a single epoch under the same optimization settings, and record (i) the wall-clock training time per epoch and (ii) the peak CUDA memory usage. Fig.~\ref{fig:Practical_Computational_Cost} summarizes the results. Both the epoch time and the peak GPU memory increase approximately linearly with $P$, which is consistent with the theoretical per-layer complexity of HGNAN being linear in the feature dimension.

\section{Conclusion}\label{conclusion}
We introduce HGNAN, an inherently interpretable neural additive architecture for learning on higher-order networks. HGNAN replaces entangled message-passing representations with a decomposable model in which feature-wise shape functions characterize nonlinear attribute effects and structural components determine how these effects are aggregated through higher-order relations. HGNAN-node provides feature- and neighbor-level interpretations for node prediction, while HGNAN-edge provides feature- and distance-level interpretations for hyperedge prediction over \(s\)-intersection graphs. Empirically, HGNAN-node remains competitive with leading HGNN baselines, while HGNAN-edge achieves strong performance on missing-reaction recovery in genome-scale metabolic networks. Cross-network transfer experiments further show that the learned feature and distance-domain responses capture reusable patterns across related metabolic systems, and case studies show that the decompositions provide biologically meaningful evidence for predictions. This dual capability of delivering strong predictive performance while offering transparent decision-making makes HGNAN a valuable tool for scientific discovery in hypergraph-based applications, especially in fields such as biotechnology, bioinformatics, and social network analysis. One limitation, however, is that HGNAN constructs a separate neural network for each feature, which can lead to high computational costs on datasets with very large dimensionality. Fortunately, such extreme high-dimensional datasets are relatively uncommon in many real-world applications. Looking forward, future work will explore extensions to dynamic hypergraphs, further broadening its applicability and scalability.

\bibliographystyle{IEEEtran}
\bibliography{references}

\appendices

\section{Additional Details for datasets} \label{dataset_info}

\textbf{Node Classification.} For Mushroom, Zoo, NTU2012, Cora, Pokec, and Actor, nodes represent samples, objects, accounts, users, or film-industry individuals, while hyperedges encode shared attributes, co-citation, co-creation relations, friendship groups, or film-level collaborations~\cite{mushroom_73,zoo_111,chen2003visual}. These datasets cover binary and multi-class node classification tasks, including edibility, animal type, 3D object category, explicit-content presence, gender, and production role prediction. Their features include categorical sample attributes, visual descriptors, account/profile metadata, and Wikipedia keyword-based attributes. Detailed statistics are reported in Table~\ref{tab:node_classification_stats}.

\begin{table}[t]
\centering
\caption{Summary of node classification datasets. $|E|$ stands for the
number of nodes on a hyperedge $E$, and $d_v$ stands for the number of
hyperedges containing a node $v$.}
\label{tab:node_classification_stats}
\scriptsize
\setlength{\tabcolsep}{3pt} 

\resizebox{\linewidth}{!}{%
\begin{tabular}{lcccccc}
\toprule
 & \textbf{NTU2012}
 & \textbf{Mushroom}
 & \textbf{Zoo}
 & \textbf{Cora}
 & \textbf{Pokec}
 & \textbf{Actor} \\
\midrule
$|V|$                  & 2012 & 8124 & 101  & 2708 & 14998 & 16255 \\
$|E|$                  & 2012 & 298  & 43   & 1579 & 2406 & 10164 \\
\# feature             & 100  & 22   & 16   & 1433 & 65   & 50    \\
\# class               & 67   & 2    & 7    & 7    & 2    & 3     \\
$\max |E|$             & 5    & 5 & 93   & 5    & 7    & 28    \\
$\min |E|$             & 5    & 1    & 2    & 2    & 2    & 1     \\
$\text{avg } |E|$      & 5.0  & 136.3 & 39.9 & 3.03 & 2.3  & 5.3   \\
$\max d_v$             & 19    & 1808 & 17   & 145  & 7    & 205   \\
$\min d_v$             & 1    & 1    & 1    & 0    & 1    & 1     \\
$\text{avg } d_v$      & 2.2  & 1.8  & 27.2 & 1.77 & 1.7  & 3.4   \\
\bottomrule
\end{tabular}%
}
\end{table}
\begin{figure*}[htbp]
    \centering
    \includegraphics[width=0.9\linewidth]{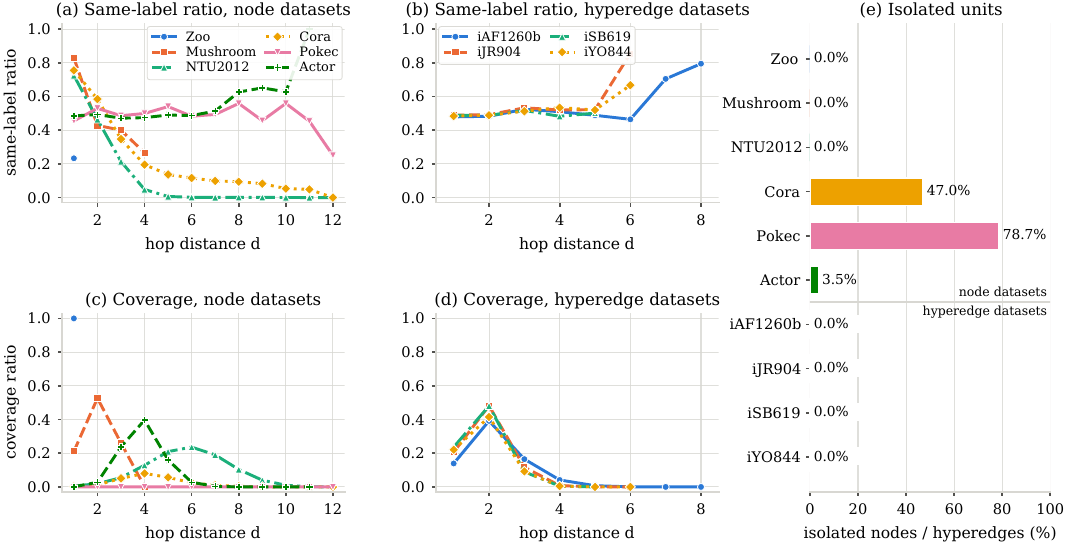}
    \caption{Distance-dependent label structure of the ten networks.
    (a)--(b) Same-label ratio $SLR(d)$ and (c)--(d) coverage $\mathrm{Cov}(d)$ as a function of
    hop distance, shown separately for the node-classification datasets and the hyperedge
    (GEM) datasets. (e) Fraction of units that have no neighbor at any finite distance.
    Both $SLR(d)$ and $\mathrm{Cov}(d)$ are averages over the units that do have a neighbor at
    distance $d$, so isolated units leave no trace in (a)--(d); panel (e) reports them
    explicitly. For the GEM datasets the distances are those used by the model, i.e.\ computed
    from the $1$-intersection graph built from training positive hyperedges only.}
    \label{fig:homophily}
\end{figure*}

\textbf{Hyperedge Prediction.} The BiGG (Biochemical Genetic and Genomic) dataset is a repository of genome-scale metabolic models (GEMs) for various organisms, including bacteria, yeast, and human cells. The BiGG dataset consists of multiple components. First, it includes genome-scale metabolic models that represent the biochemical processes of various organisms. These models define metabolic reactions, describing the biochemical transformations occurring within cells, and metabolites, which are the chemical compounds involved in these reactions.

\begin{table}[t]
\centering
\caption{Summary of four datasets from the BiGG dataset.}
\label{tab:comparison_stats}
\begin{tabular}{lcccc}
\toprule
                         & \textbf{iAF1260b} & \textbf{iJR904} & \textbf{iSB619} & \textbf{iYO844} \\
\midrule
$|V|$                   & 2388             & 1075          & 743           & 1250          \\
$|E|$                   & 1668             & 761           & 655           & 990           \\
Missing Rate          & 0.236
             & 0.088
          & 0.171
         & 0.160
           \\
$\max |E|$               & 67               & 56            & 61            & 63            \\
$\min |E|$               & 1                & 1             & 1             & 1             \\
$\text{avg } |E|$        & 3.88             & 4.18          & 5.14          & 4.19          \\
$\max d_v$                 & 912              & 259           & 362           & 616           \\
$\min d_v$                 & 1                & 1             & 1             & 1             \\
$\text{avg } d_v$          & 5.55             & 4.77          & 5.83          & 5.29          \\
\bottomrule
\end{tabular}
\end{table}
\textbf{Feature generation.} Since BiGG models do not provide metabolite features, we construct interpretable molecular features using MACCS Keys. Unlike node2vec embeddings or generic SMILES-based fingerprints, MACCS Keys encode each molecule as a 167-dimensional binary vector, where each entry indicates the presence of a predefined substructure or functional group. Because these substructures are chemically meaningful, the resulting features are more interpretable. For metabolites without valid MACCS-key features, we use zero padding and report the missing-feature rate in Table~\ref{tab:comparison_stats}.

\textbf{Access.} The BiGG dataset is publicly available at \url{http://bigg.ucsd.edu/}. MACCS Keys are available at \url{https://github.com/jAniceto/ml-knowledge-base/blob/main/02-data-preparation/feature-engineering/maccs.md}.
Table \ref{tab:comparison_stats} provides details about the four models from the BiGG dataset used in our paper.

\textbf{Homophily.}
To quantify homophily at different distances $d$ on a hypergraph, we report the same-label ratio:
\begin{equation*}
    SLR(d) =  \frac{1}{|V_d|}\frac{\sum_{j=1}^N \mathbf{1}(y_j=y_i) \mathbf{1}(d(i,j)=d)}{\sum_{j=1}^N \mathbf{1}(d(i,j)=d)}
\end{equation*}
where $V_d=\{ i: \exists j\neq i, d(i,j)=d \} \}$ is the set of nodes that have at least one neighbor exactly $d$ hops away. Since $SLR(d)$ can be misleading when the number of nodes at that distance is small, which is common when distance is large, we also provide coverage ratio across distances for reference: 
\[
    \mathrm{Cov}(d)
    =\frac{1}{N}\sum_{i=1}^{N}
    \frac{
    \sum_{j\neq i}
    \mathbf{1}(d(i,j)=d)
    }{
    N-1
    }.
\]

Fig. \ref{fig:homophily} shows that the label structure is strongly distance-dependent and differs sharply between the two groups of networks. Among the node-classification datasets, Mushroom, NTU2012 and Cora are locally homophilic, with $SLR(1)$ between $0.75$ and $0.82$ decaying towards zero within a few hops, whereas Pokec stays near $0.5$ at every distance and Actor increases with distance. The hyperedge datasets instead start near $0.5$ and rise at large distances, reaching $0.67$-$0.79$ beyond six hops.

Panel (e) reports a property that the two curves cannot show. Because $SLR(d)$ and $\mathrm{Cov}(d)$ average over the units that have at least one neighbor at distance $d$, units with no neighbor at all are silently excluded: $47.0\%$ of Cora nodes and $78.7\%$ of Pokec nodes fall in this category, as do $5.9\%$-$9.2\%$ of the candidate hyperedges in the GEM datasets, while Zoo, Mushroom and NTU2012 have none. For these units a hypergraph model has no structural evidence to aggregate and the prediction necessarily falls back on the attributes of the unit itself, which is consistent with the limited effect of removing aggregation on Pokec reported in Section~\ref{sec:ablation}. Taken together, the panels indicate that the datasets do not follow a uniform homophily assumption and motivate distance-aware structural modeling that does not presuppose informative local neighborhoods.

\section{Details for Model Training}
\label{training_info}

All datasets are randomly split into training, validation, and test sets with a 2:1:1 ratio. Each experiment is repeated over 10 random seeds using the same splits and seeds for all methods, and we report the mean and standard deviation of test performance. All experiments are run on an NVIDIA V100 GPU. For node classification, we tune all methods over learning rate $\{0.01,0.001\}$, weight decay $\{0,0.0005\}$, and hidden channels $\{64,128,256\}$. For attention-based methods, we additionally tune the number of heads in $\{1,4,8\}$; for ED-HNN, the number of layers in $\{0,1\}$ and restart rate in $\{0,0.5\}$; and for HGNAN-node, the number of layers in $\{3,5\}$, dropout in $\{0,0.5\}$, and $s_{\max}\in\{1,2,3\}$. We select the configuration with the best aggregated validation accuracy across the four datasets, and also report dataset-specific best configurations following \cite{Chien2021YouAA,wang2022equivariant}. 

For hyperedge prediction, for each split, the \(s\)-intersection graph and all structural distances used by the model are constructed exclusively from the positive hyperedges in the training set. Validation and test positive hyperedges are excluded from structural graph construction. All hyperedge-prediction methods are tuned over learning rate $\{0.001,0.01\}$ and weight decay $\{0,0.0005\}$. We additionally tune HGNAN-edge over dropout $\{0,0.5\}$, hidden channels $\{32,64,128\}$, and $s_{\max}\in\{1,2,3\}$; CHESHIRE over embedding dimension $\{128,256\}$, convolutional dimension $\{64,128\}$, Chebyshev order $\{3,5\}$, and dropout $\{0.1,0.2\}$; NHP over embedding and convolutional dimensions $\{128,256\}$ and $\{64,128\}$; and HyperSAGNN over embedding dimension $\{128,256\}$, convolutional dimension $\{64,128\}$, and attention heads $\{1,3\}$.

\section{Hyperedge Prediction with Node-Level Molecular Features}
\label{app:edge_maccs}

In Table~\ref{tab:edge_result} the baselines operate on the incidence structure alone, as in their original
implementations, so part of the reported gap may simply reflect the fact that HGNAN-edge additionally sees MACCS-derived reaction features. To separate the two effects, we give every baseline the same node-level MACCS fingerprints and re-tune it under identical splits and protocol. The 167-dimensional signed MACCS differences provide explicit reaction-level semantics but can discard node-level information through aggregation, whereas the baselines receive metabolite fingerprints for each node. We therefore augment this representation with fingerprint means and standard deviations, Tanimoto-similarity statistics, and hyperedge size, yielding 504 features derived from the same molecular inputs.
Table~\ref{tab:app_edge_maccs} shows that molecular features account for a part of the gap and that this richer representation substantially improves performance while preserving additivity in the constructed hyperedge features. Once they are available, CHESHIRE becomes stronger than the $167$-dimensional HGNAN-edge in AUROC on iAF1260b, iJR904 and iYO844, whereas the $504$-dimensional variant is the best or statistically indistinguishable from the best method on every dataset and metric, and no baseline is significantly better than it anywhere. The difference after changing pooling indicates that HGNAN-edge can be more powerful if it receives more feature information via diverse pooling. 

\begin{table}[t]
\centering
\caption{Hyperedge prediction when all baselines additionally receive node-level MACCS fingerprints. Entries are mean $\pm$ standard deviation (\%) over 10 runs with identical splits for all methods. Bold marks the best entry in each column; $\dagger$ marks baseline entries whose difference from HGNAN-edge (504-d) is not statistically significant (paired Wilcoxon, $p\ge0.05$), and all unmarked baseline entries are significantly worse.}
\label{tab:app_edge_maccs}
\setlength{\tabcolsep}{4pt}
\resizebox{\linewidth}{!}{
\begin{tabular}{llcccc}
\toprule
Metric & Method & iAF1260b & iJR904 & iSB619 & iYO844 \\
\midrule
\multirow{5}{*}{AUROC} & CHESHIRE & $\mathbf{95.76 \pm 0.53}^{\dagger}$ & $94.53 \pm 1.05$ & $90.71 \pm 1.16$ & $95.16 \pm 0.91$ \\
 & NHP & $89.34 \pm 0.63$ & $84.26 \pm 1.43$ & $81.18 \pm 2.24$ & $87.98 \pm 1.14$ \\
 & HyperSAGNN & $94.36 \pm 0.48$ & $92.19 \pm 1.22$ & $86.51 \pm 1.07$ & $93.43 \pm 0.93$ \\
 & HGNAN-edge (167-d) & $91.42 \pm 0.97$ & $90.81 \pm 2.07$ & $90.42 \pm 1.15$ & $90.80 \pm 1.25$ \\
 & HGNAN-edge (504-d) & $95.34 \pm 0.77$ & $\mathbf{95.51 \pm 0.51}$ & $\mathbf{94.45 \pm 0.88}$ & $\mathbf{96.67 \pm 0.52}$ \\
\midrule
\multirow{5}{*}{Accuracy} & CHESHIRE & $\mathbf{89.31 \pm 1.13}^{\dagger}$ & $86.32 \pm 1.68$ & $81.79 \pm 2.77$ & $87.52 \pm 2.56$ \\
 & NHP & $77.48 \pm 0.71$ & $72.42 \pm 1.15$ & $71.88 \pm 1.63$ & $74.50 \pm 1.15$ \\
 & HyperSAGNN & $87.55 \pm 1.21$ & $84.76 \pm 0.87$ & $78.73 \pm 1.63$ & $86.00 \pm 1.70$ \\
 & HGNAN-edge (167-d) & $85.80 \pm 0.83$ & $86.60 \pm 2.02$ & $85.01 \pm 2.80$ & $85.86 \pm 1.28$ \\
 & HGNAN-edge (504-d) & $89.27 \pm 0.97$ & $\mathbf{90.99 \pm 0.66}$ & $\mathbf{88.78 \pm 1.19}$ & $\mathbf{91.25 \pm 0.96}$ \\
\midrule
\multirow{5}{*}{AUPRC} & CHESHIRE & $\mathbf{94.84 \pm 0.80}^{\dagger}$ & $\mathbf{94.14 \pm 1.20}^{\dagger}$ & $90.19 \pm 1.44$ & $95.16 \pm 1.10$ \\
 & NHP & $89.53 \pm 0.83$ & $83.59 \pm 0.92$ & $78.68 \pm 3.67$ & $88.12 \pm 1.34$ \\
 & HyperSAGNN & $94.28 \pm 0.51^{\dagger}$ & $92.09 \pm 1.36$ & $87.02 \pm 2.09$ & $93.88 \pm 1.06$ \\
 & HGNAN-edge (167-d) & $89.53 \pm 1.58$ & $87.54 \pm 3.12$ & $87.40 \pm 2.08$ & $88.35 \pm 2.29$ \\
 & HGNAN-edge (504-d) & $94.56 \pm 1.14$ & $94.10 \pm 0.97$ & $\mathbf{93.45 \pm 1.57}$ & $\mathbf{96.48 \pm 0.78}$ \\
\bottomrule
\end{tabular}
}
\end{table}

\section{Additional Faithfulness Evaluation}
\label{app:faithfulness}

We provide additional quantitative evaluation of the faithfulness of the feature- and structure-level explanations produced by HGNAN. Following the comprehensiveness and sufficiency framework
\cite{deyoung2020eraser}, \emph{comprehensiveness} measures how much the prediction deteriorates after removing the components identified as most important, whereas \emph{sufficiency} measures how much the prediction
deteriorates when only those components are retained. We denote the two metrics at a perturbation budget $q$ by COMP@$q$ and SUFF@$q$, respectively. Higher COMP and lower SUFF indicate more faithful attribution. Feature-level explanations are compared with Gradient$\times$Input (G$\times$I) \cite{ancona2017towards} and random
attribution.

For node prediction, feature removal replaces the selected features with their training-set means, and prediction degradation is measured by the decrease in the probability of the original predicted class. For hyperedge prediction, the selected reaction-feature basis values are set to zero, corresponding to no net change in the associated functional group, and degradation is measured by the decrease in AUROC. For node-level structural evaluation, learned neighbor weights are treated as structural-importance scores and are renormalized after neighbor removal. In contrast, HGNAN-edge admits an exact additive decomposition into source-hyperedge terms $\kappa^{(s)}(e_i,e_j)\phi(e_j)$, so structural contributions can be removed directly without renormalization.

\subsection{Node-Level Feature Faithfulness}

Table~\ref{tab:app_node_feature_faithfulness} summarizes feature-level faithfulness at a 20\% perturbation budget. HGNAN outperforms random attribution by a wide margin on all three datasets, and outperforms G$\times$I on NTU2012 and Zoo. On Pokec, where the aggregated structural term contributes little, the margin against G$\times$I is smaller. Table~\ref{tab:app_ntu_feature_curve} further reports the full perturbation curve on NTU2012, where HGNAN achieves higher comprehensiveness and lower sufficiency than G$\times$I over nearly the entire range of perturbation budgets. Together, these results indicate that HGNAN-node provides faithful feature-level explanations on the datasets where its predictions are themselves informative.

\begin{table*}[t]
\centering
\caption{Node-level feature attribution faithfulness at a 20\% perturbation budget. Higher COMP and lower SUFF indicate more faithful attribution. Results are averaged over 10 runs. HGNAN outperforms G$\times$I and random attribution on all three datasets.}
\label{tab:app_node_feature_faithfulness}
\setlength{\tabcolsep}{7pt}
\begin{tabular}{lcccccc}
\toprule
& \multicolumn{2}{c}{HGNAN}
& \multicolumn{2}{c}{Gradient$\times$Input}
& \multicolumn{2}{c}{Random} \\
\cmidrule(lr){2-3}
\cmidrule(lr){4-5}
\cmidrule(lr){6-7}
Dataset
& COMP@20\% $\uparrow$& SUFF@20\%$\downarrow$
& COMP@20\%$\uparrow$ & SUFF@20\%$\downarrow$
& COMP@20\%$\uparrow$ & SUFF@20\% $\downarrow$\\
\midrule
NTU2012
& $\mathbf{0.5019\pm0.0190}$
& $\mathbf{0.2324\pm0.0196}$
& $0.3610\pm0.0123$
& $0.3912\pm0.0178$
& $0.0741\pm0.0082$
& $0.7656\pm0.0058$ \\
Zoo
& $\mathbf{0.2700\pm0.0724}$
& $\mathbf{0.2589\pm0.0576}$
& $0.2367\pm0.0659$
& $0.2980\pm0.0425$
& $0.0538\pm0.0226$
& $0.4970\pm0.0352$ \\
Pokec
& $\mathbf{0.0997\pm0.0063}$
& $\mathbf{0.0069\pm0.0018}$
& $0.0916\pm0.0066$
& $0.0141\pm0.0049$
& $0.0202\pm0.0012$
& $0.0841\pm0.0046$ \\
\bottomrule
\end{tabular}
\end{table*}

\begin{table*}[t]
\centering
\caption{Full feature-attribution faithfulness curve on NTU2012. COMP removes the top-$q$ fraction of attributed features, whereas SUFF retains only the top-$q$ fraction. Results are mean $\pm$ standard deviation over 10 runs. HGNAN achieves higher comprehensiveness and lower sufficiency than G$\times$I over nearly the entire range of perturbation budgets}
\label{tab:app_ntu_feature_curve}
\setlength{\tabcolsep}{6pt}
\begin{tabular}{clccc}
\toprule
Metric & $q$ & HGNAN & Gradient$\times$Input & Random \\
\midrule
\multirow{9}{*}{COMP$\uparrow$}
& 0.10 & $\mathbf{0.2572\pm0.0175}$ & $0.2090\pm0.0108$ & $0.0301\pm0.0045$ \\
& 0.20 & $\mathbf{0.5019\pm0.0190}$ & $0.3610\pm0.0123$ & $0.0741\pm0.0082$ \\
& 0.30 & $\mathbf{0.6850\pm0.0136}$ & $0.4715\pm0.0159$ & $0.1352\pm0.0142$ \\
& 0.40 & $\mathbf{0.7857\pm0.0094}$ & $0.5554\pm0.0135$ & $0.2246\pm0.0180$ \\
& 0.50 & $\mathbf{0.8318\pm0.0092}$ & $0.6358\pm0.0135$ & $0.3483\pm0.0189$ \\
& 0.60 & $\mathbf{0.8522\pm0.0094}$ & $0.7167\pm0.0091$ & $0.5003\pm0.0159$ \\
& 0.70 & $\mathbf{0.8619\pm0.0097}$ & $0.7827\pm0.0068$ & $0.6493\pm0.0105$ \\
& 0.80 & $\mathbf{0.8668\pm0.0097}$ & $0.8289\pm0.0070$ & $0.7670\pm0.0053$ \\
& 0.90 & $\mathbf{0.8696\pm0.0097}$ & $0.8568\pm0.0088$ & $0.8380\pm0.0078$ \\
\midrule
\multirow{9}{*}{SUFF$\downarrow$}
& 0.10 & $\mathbf{0.5084\pm0.0168}$ & $0.5937\pm0.0138$ & $0.8373\pm0.0084$ \\
& 0.20 & $\mathbf{0.2324\pm0.0196}$ & $0.3912\pm0.0178$ & $0.7656\pm0.0058$ \\
& 0.30 & $\mathbf{0.1054\pm0.0136}$ & $0.2856\pm0.0145$ & $0.6470\pm0.0102$ \\
& 0.40 & $\mathbf{0.0482\pm0.0084}$ & $0.2145\pm0.0133$ & $0.4980\pm0.0153$ \\
& 0.50 & $\mathbf{0.0206\pm0.0069}$ & $0.1523\pm0.0131$ & $0.3463\pm0.0162$ \\
& 0.60 & $\mathbf{0.0074\pm0.0046}$ & $0.0933\pm0.0113$ & $0.2215\pm0.0130$ \\
& 0.70 & $\mathbf{0.0019\pm0.0025}$ & $0.0463\pm0.0061$ & $0.1340\pm0.0109$ \\
& 0.80 & $\mathbf{-0.0002\pm0.0021}$ & $0.0143\pm0.0035$ & $0.0733\pm0.0078$ \\
& 0.90 & $-0.0006\pm0.0014$ & $\mathbf{-0.0020\pm0.0021}$ & $0.0311\pm0.0047$ \\
\bottomrule
\end{tabular}
\end{table*}

\subsection{Node-Level Structural Faithfulness}

We next evaluate whether the sparse adaptive aggregation weights correctly identify structurally influential neighbors. Table~\ref{tab:app_node_neighbor} reports the probability decrease obtained by removing the neighbors with the largest learned weights and compares it with random removal. On NTU2012, removing the highest-weight neighbor causes a $12.2\times$ larger probability decrease than random removal. When changing the number of top important neighbors removed, the ratio decreases to $3.0\times$ after removing the top ten neighbors, since ten neighbors already account for most of the achievable effect. The effect is weaker on Pokec but still significant.

\begin{table}[t]
\centering
\caption{Node-level structural faithfulness under neighbor removal. $\Delta$Prob denotes the decrease in the probability of the original predicted class. Ratio denotes the degradation under importance-based
removal divided by random removal. Results are mean $\pm$ standard deviation over 10 runs. Removing the highest-weight neighbor causes a significantly larger probability decrease than random removal.}
\label{tab:app_node_neighbor}
\setlength{\tabcolsep}{6pt}
\begin{tabular}{llccc}
\toprule
Dataset & $k$
& Importance $\Delta$Prob
& Random $\Delta$Prob
& Ratio
\\
\midrule
\multirow{4}{*}{NTU2012}
& 1   & $0.0048\pm0.0021$ & $0.0004\pm0.0002$ & $\mathbf{12.2\times}$ \\
& 3   & $0.0160\pm0.0027$ & $0.0017\pm0.0006$ & $\mathbf{9.7\times}$ \\
& 10  & $0.0387\pm0.0034$ & $0.0128\pm0.0020$ & $\mathbf{3.0\times}$ \\
& 20  & $0.0419\pm0.0038$ & $0.0248\pm0.0027$ & $1.7\times$ \\
\midrule
\multirow{4}{*}{Pokec}
& 1   & $0.0014\pm0.0003$ & $0.0008\pm0.0001$ & $1.8\times$ \\
& 3   & $0.0015\pm0.0003$ & $0.0011\pm0.0002$ & $1.4\times$ \\
& 10  & $0.0016\pm0.0003$ & $0.0015\pm0.0002$ & $1.1\times$ \\
& 20  & $0.0016\pm0.0003$ & $0.0016\pm0.0003$ & $1.0\times$ \\
\bottomrule
\end{tabular}
\end{table}

\subsection{Hyperedge-Level Feature Faithfulness}

Table~\ref{tab:app_edge_feature} reports feature comprehensiveness at a 20\% perturbation budget across the four metabolic networks and two values of $s_{\max}$. HGNAN achieves higher COMP than G$\times$I on iAF1260b and iSB619, whereas G$\times$I is stronger on iJR904 and iYO844. Nevertheless, both methods substantially outperform random attribution. Across all eight dataset--$s_{\max}$ settings, HGNAN also achieves equal or lower SUFF than G$\times$I, indicating that the features identified by HGNAN form compact predictive subsets.

\begin{table}[t]
\centering
\caption{Hyperedge-level feature comprehensiveness at a 20\% perturbation budget. Entries report the decrease in AUROC after removing the top 20\% attributed features. Higher values indicate greater comprehensiveness.}
\label{tab:app_edge_feature}
\setlength{\tabcolsep}{7pt}
\begin{tabular}{lccc}
\toprule
Dataset
& HGNAN
& Gradient$\times$Input
& Random \\
\midrule
iAF1260b
& \textbf{30.4}
& 25.3
& 3.0 \\

iJR904
& 29.5
& \textbf{36.4}
& 2.4 \\

iSB619
& \textbf{16.9}
& 5.4
& 1.0 \\

iYO844
& 26.8
& \textbf{35.5}
& 1.7 \\

\bottomrule
\end{tabular}
\end{table}

\section{Hyperedge-Level Structural Faithfulness}

HGNAN-edge provides a stronger structural faithfulness test because its
prediction can be written as an exact sum of source-hyperedge
contributions. We first separate the target hyperedge's self term
($j=i$) from the non-self structural context ($j\neq i$).
Table~\ref{tab:app_edge_self_context} shows that removing the self term
causes the largest degradation, but non-self context also contributes
substantially. Removing all structural context reduces
AUROC by 7.17 points on iAF1260b and 6.26 points on iSB619.

\begin{table}[t]
\centering
\caption{Decomposition of HGNAN-edge predictions into target-hyperedge self contribution and non-self structural context. Entries report $\Delta$AUROC after removing the corresponding additive terms. Context/self is the ratio of the full contextual effect to the self effect. Results are mean $\pm$ standard deviation over 10 runs.}
\label{tab:app_edge_self_context}
\setlength{\tabcolsep}{6pt}
\resizebox{\linewidth}{!}{
\begin{tabular}{lccc}
\toprule
 Dataset
& Remove self
& Remove all context
& Context/self
\\
\midrule
 iAF1260b & $26.98\pm1.86$ & $7.17\pm0.59$ & 26.6\% \\
 iJR904 & $30.85\pm3.89$ & $2.92\pm1.39$ & 9.5\% \\
 iSB619 & $29.60\pm2.63$ & $6.26\pm1.34$ & 21.1\% \\
 iYO844 & $30.29\pm2.97$ & $3.38\pm1.08$ & 11.2\% \\
\bottomrule
\end{tabular}
}
\end{table}

We further test whether the largest-magnitude non-self contributions
identify the small subset of source hyperedges responsible for the
contextual effect. Table~\ref{tab:app_edge_structural_topk} compares
removing the top-$k$ exact additive contributions with random source
removal. On iAF1260b and iSB619, removing only the top ten source
hyperedges decreases AUROC by 4.70 and 5.04 points, respectively, while
randomly removing ten sources changes AUROC by at most 0.02 points.
These top ten contributions account for approximately 66\% and 81\% of
the full contextual effects on the two datasets, showing that the explanations given by HGNAN-edge are faithful.

\begin{table*}[t]
\centering
\caption{Structural faithfulness of HGNAN-edge under source-hyperedge
removal. Non-self source hyperedges are ranked by
$|\kappa^{(s)}(e_i,e_j)\phi(e_j)|$. Entries report $\Delta$AUROC after
removing the top-$k$ ranked sources or $k$ randomly selected sources.}
\label{tab:app_edge_structural_topk}
\setlength{\tabcolsep}{7pt}
\begin{tabular}{llrrrrrr}
\toprule
Dataset & Removal
& $k=1$ & $k=3$ & $k=5$ & $k=10$ & $k=25$ & $k=50$ \\
\midrule
\multirow{2}{*}{iAF1260b}
& HGNAN & 0.97 & 2.78 & 3.76 & 4.70 & 4.86 & \textbf{4.89} \\
& Random & 0.00 & -0.00 & -0.00 & 0.00 & 0.00 & 0.01 \\
\midrule
\multirow{2}{*}{iJR904}
& HGNAN & 0.20 & 0.38 & 0.46 & 0.51 & 0.62 & \textbf{0.73} \\
& Random & 0.00 & 0.00 & 0.01 & 0.00 & 0.01 & 0.01 \\
\midrule
\multirow{2}{*}{iSB619}
& HGNAN & 0.22 & 1.22 & 2.63 & \textbf{5.04} & 4.57 & 4.94 \\
& Random & -0.00 & 0.00 & 0.01 & 0.02 & 0.04 & 0.03 \\
\midrule
\multirow{2}{*}{iYO844}
& HGNAN & -0.10 & 0.64 & 0.47 & 0.59 & 0.89 & \textbf{1.00} \\
& Random & -0.00 & 0.00 & -0.00 & 0.01 & 0.06 & 0.08 \\
\bottomrule
\end{tabular}
\end{table*}

\section{Additional Results for Cross-Dataset Transfer}
\label{app:transfer}

We provide the complete results for the zero-shot cross-dataset transfer
experiment described in Section~\ref{sec:cross_bigg_transfer}. For each source-target pair, all model parameters are learned on the source GEM and frozen when the model is evaluated on the target GEM. No target-domain fine-tuning is performed. This setting therefore directly evaluates whether the mechanisms learned on one metabolic network can be reused on another.

\begin{table*}[t]
    \centering
    \caption{Zero-shot cross-dataset transfer results across the four BiGG GEM datasets. AUROC is used as the evaluation metric. For each source-target
    pair, all model parameters are trained on the source GEM and frozen during evaluation on the target GEM, with no target-domain fine-tuning. Bold
    values indicate the best result for each transfer direction. Results are mean $\pm$ standard deviation over 10 runs.}
    \label{tab:crossdataset}
    \begin{tabular}{cccccc}
        \toprule
        \textbf{Training Data} & \textbf{Testing Data} &
        \textbf{NHP} & \textbf{HyperSAGNN} &
        \textbf{CHESHIRE} & \textbf{HGNAN-edge} \\
        \midrule
        iAF1260b & iJR904
        & $0.6213\pm0.0468$ & $0.7769\pm0.0204$
        & $0.6785\pm0.0440$ & $\mathbf{0.8992\pm0.0183}$ \\
        iAF1260b & iSB619
        & $0.6460\pm0.0273$ & $\mathbf{0.7445\pm0.0391}$
        & $0.5455\pm0.0593$ & $0.5497\pm0.0140$ \\
        iAF1260b & iYO844
        & $0.6284\pm0.0397$ & $\mathbf{0.7479\pm0.0250}$
        & $0.5059\pm0.0390$ & $0.7068\pm0.0590$ \\
        iJR904 & iAF1260b
        & $0.6476\pm0.0119$ & $0.7657\pm0.0031$
        & $0.6717\pm0.0257$ & $\mathbf{0.8555\pm0.0154}$ \\
        iJR904 & iSB619
        & $0.6291\pm0.0295$ & $\mathbf{0.7137\pm0.0430}$
        & $0.5017\pm0.0316$ & $0.5477\pm0.0726$ \\
        iJR904 & iYO844
        & $0.6667\pm0.0377$ & $0.7306\pm0.0674$ 
        & $0.6406\pm0.0158$ & $\mathbf{0.7641\pm0.0331}$\\
        iSB619 & iAF1260b
        & $0.6754\pm0.0472$ & $0.7333\pm0.0346$
        & $0.5827\pm0.0450$ & $\mathbf{0.8072\pm0.0139}$ \\
        iSB619 & iJR904
        & $0.5798\pm0.0042$ & $0.8332\pm0.0422$
        & $0.5981\pm0.0186$ & $\mathbf{0.8561\pm0.0153}$ \\
        iSB619 & iYO844
        & $0.6598\pm0.0215$ & $0.8218\pm0.0552$
        & $0.6404\pm0.0190$ & $\mathbf{0.8351\pm0.0208}$ \\
        iYO844 & iAF1260b
        & $0.6551\pm0.0108$ & $0.7627\pm0.0477$
        & $0.6081\pm0.0562$ & $\mathbf{0.8415\pm0.0092}$ \\
        iYO844 & iJR904
        & $0.6143\pm0.0523$ & $0.8191\pm0.0156$
        & $0.6637\pm0.0534$ & $\mathbf{0.8967\pm0.0130}$ \\
        iYO844 & iSB619
        & $0.6420\pm0.0255$ & $\mathbf{0.7950\pm0.0577}$
        & $0.5888\pm0.0245$ & $0.5193\pm0.0202$ \\
        \bottomrule
    \end{tabular}
\end{table*}

Table~\ref{tab:crossdataset} shows a clear target-dependent transfer pattern. HGNAN-edge achieves the best AUROC in 8 of the 12 transfer directions,
including all three transfers to iAF1260b, all three transfers to iJR904, and one of the three transfers to iYO844. In contrast, it does not achieve the best result in any of the three transfers to iSB619. Excluding iSB619 as the target, HGNAN-edge achieves an average AUROC of $0.825$, compared with $0.781$ for HyperSAGNN, $0.639$ for NHP, and $0.621$ for CHESHIRE, and is the best-performing method in 8 of the remaining 9 transfer directions.

The weaker transfer to iSB619 is closely associated with the feature distribution used by HGNAN-edge. Unlike the compared baselines, which use only the incidence structure, HGNAN-edge additionally uses MACCS-derived reaction features. The distribution of MACCS key's values of nodes from iSB619 contains substantially more extreme values than those of the other GEMs, producing a pronounced covariate shift when a model trained on another network is
transferred to iSB619 without adaptation. The incidence-only baselines are not exposed to this additional source of distribution shift. Consequently,
the lower HGNAN-edge performance on iSB619 should not be interpreted solely as a failure of structural transfer, since zero-shot performance reflects
both transfer of the learned structural mechanism and alignment of the reaction-feature distributions. Overall, these results indicate that HGNAN-edge can reuse its learned feature-wise and distance-aware structural functions across related metabolic networks
without target-domain adaptation, while also revealing feature-distribution shift as an important boundary condition for zero-shot transfer.

\end{document}